\documentclass[times,referee,twocolumn,final,authoryear]{elsarticle}

\usepackage{ycviu}
\usepackage{framed,multirow}

\usepackage{amssymb}
\usepackage{latexsym}

\usepackage{url}
\usepackage{xcolor}
\definecolor{newcolor}{rgb}{.8,.349,.1}

\usepackage{enumerate}
\usepackage[shortlabels]{enumitem}
\usepackage{cleveref}
\usepackage{amssymb}
\usepackage{amsmath}
\usepackage{float}
\usepackage{times}
\usepackage{amsmath,amsthm}
\usepackage{tikz}
\usepackage{wrapfig}
\usepackage{algorithm}
\usepackage{algorithmic}
\usepackage{multirow}
\usepackage{subfig}
\usepackage{graphicx}
\usepackage{epstopdf}
\usepackage{color}
\usepackage{cleveref}
\usepackage{amssymb}
\usepackage{array}
\newcolumntype{L}[1]{>{\raggedright\let\newline\\\arraybackslash\hspace{0pt}}m{#1}}
\newcolumntype{C}[1]{>{\centering\let\newline\\\arraybackslash\hspace{0pt}}m{#1}}
\newcolumntype{R}[1]{>{\raggedleft\let\newline\\\arraybackslash\hspace{0pt}}m{#1}}

\usepackage{latexsym}
\allowdisplaybreaks
\usepackage{xspace}
\usepackage[draft]{fixme}
\fxsetup{layout=inline,theme=color}
\definecolor{fxnote}{rgb}{0.8000,0.0000,0.0000}
\colorlet{fxnotebg}{yellow}
\makeatletter
\renewcommand*\FXLayoutInline[3]{%
	\@fxdocolon {#3}{%
		\@fxuseface {inline}%
		\colorbox{fx#1bg}{\color {fx#1}\ignorespaces #3\@fxcolon #2}}}
\makeatother

\makeatletter
\DeclareRobustCommand\onedot{\futurelet\@let@token\@onedot}
\def\@onedot{\ifx\@let@token.\else.\null\fi\xspace}
 
\def\ie{\emph{i.e}\onedot}

\def\wrt{w.r.t\onedot} 
\def\etal{\emph{et al}\onedot}
\makeatother

\journal{Computer Vision and Image Understanding}

\begin{document}

\begin{frontmatter}

\title{Simple to Complex Cross-modal Learning to Rank}

\author[1]{Minnan \snm{Luo}}
\ead{minnluo@xjtu.edu.cn}
\author[2]{Xiaojun \snm{Chang}\corref{cor1}}
\ead{cxj273@gmail.com}
\cortext[cor1]{Corresponding author.}
\author[3]{Zhihui \snm{Li}}
\ead{zhihuilics@gmail.com}
\author[4]{Liqiang \snm{Nie}}
\ead{nieliqiang@gmail.com}
\author[2]{Alexander G. \snm{Hauptmann}}
\ead{alex@cs.cmu.edu}
\author[1]{Qinghua \snm{Zheng}}
\ead{qhzheng@xjtu.edu.cn}

\address[1]{SPKLSTN Lab, Department of Computer Science, Xi'an Jiaotong University, Xi'an, China}
\address[2]{School of Computer Science, Carnegie Mellon University, PA, USA}
\address[3]{Faculty of Engineering and Information Technology, University of Technology Sydney, Australia}
\address[4]{School of Computing, National University of Singapore, Singapore}

\received{22 August 2016}
\finalform{June 22 2017}
\accepted{4 July 2017}
\availableonline{xxx}

\begin{abstract}
The heterogeneity-gap between different modalities brings a significant challenge to multimedia information retrieval. Some studies formalize the cross-modal retrieval tasks as a ranking problem and learn a shared multi-modal embedding space to measure the cross-modality similarity.
However, previous methods often establish the shared embedding space based on linear mapping functions which might not be sophisticated enough to reveal more complicated inter-modal correspondences.
Additionally, current studies assume that the rankings are of equal importance, and thus all rankings are used simultaneously, or a small number of rankings are selected randomly to train the embedding space at each iteration. Such strategies, however, always suffer from outliers as well as reduced generalization capability due to their lack of insightful understanding of procedure of human cognition.
In this paper, we involve the self-paced learning theory with diversity into the cross-modal learning to rank and learn an optimal multi-modal embedding space based on non-linear mapping functions. This strategy enhances the model's robustness to outliers and achieves better generalization via training the model gradually from easy rankings by diverse queries to more complex ones.
An efficient alternative algorithm is exploited to solve the proposed challenging problem with fast convergence in practice.
Extensive experimental results on several benchmark datasets indicate that the proposed method achieves significant improvements over the state-of-the-arts in this literature.   
\end{abstract}


\end{frontmatter}


\section{Introduction}\label{sec:introduction}
\label{sec:intro}

In many real-world applications, data related to the same underlying object (content) are often exhibited in diverse modalities for better human cognition \cite{lux2004cross,zhai2013heterogeneous}. 
For example, when we want to know what is a dinosaur, we prefer to find the results across various modalities, such as searching images (videos) to figure out what a dinosaur looks like, also searching text description on its size and other biology information for best comprehension. 
As a result, cross-modal retrieval attracts increasing attention and plays an important role to describe the content of an image with natural language and conversely retrieve image given textual query \cite{pereira2014cross,amir2004multi,ChangYYX17}. 
However, since data in diverse modalities are presented in heterogeneous feature spaces and usually have varying statistical properties, it is a significant challenge to bridge the heterogeneity-gap between multi-modal data \cite{grangier2008discriminative,RanjanRJ15}. 

In the past decades, a large number of efforts have been devoted to revealing the inter-modal correspondence via learning a shared embedding space for cross-modal similarity measurement \cite{KangLHWNXP15,WangHKXP15,JinMZZX15,MenonSC15,IrieAT15,ChangMY17}.
For example, Canonical Correlation Analysis (CCA) and its extensions to kernel version \cite{hotelling1936relations,hardoon2004canonical} aim to learn a common representation by mutually maximizing the correlation between their projections onto the shared basis vectors;
Latent Dirichlet Allocation (LDA) based methods  \cite{blei2003modeling,barnard2003matching,wang2009simultaneous,jia2011learning,ChangMLY17} establish the shared latent semantic model through the joint distribution of images and the corresponding annotations as well as the conditional relationships between them. However, these methods separate the shared space learning from the ultimate ranking performance, and thus usually suffer from poor generalization capability \cite{lu2013low}.
Motivated by the Ordered Weighted Pairwise Classification (OWPC) loss \cite{usunier2009ranking}, Weston \etal \cite{weston2011wsabie} involved a dynamic importance for different ranking and proposed a Weighted Approximate-Rank Pairwise loss (WARP) for multi-label annotation problem.
Gong \etal \cite{gong2013deep} developed the deep extension of this method. However, WARP is parameterized by a set of decreasing weights which are predefined.
Instead, Cross-modal Learning to Rank (CMLR) learns the multi-modal embedding space through minimizing a ranking-based loss function \cite{grangier2008discriminative,mcfee2010metric,DengTYLG16,YouLJY16,KangXLXP15}. 
Since CMLR is designed orientating to the performance of cross-modal ranking directly, it has become an increasingly important research direction in cross-modal information retrieval. Many approaches have been proposed based on this strategy \cite{wu2013cross,wu2015cross,zhu2013linear,jiang2015deep,wang2015joint,wang2014multi,li2013mlrank,habibian2015discovering}. 

However, the task of CMLR remains a significant challenge because it requires an understanding of the content of images, sentences, and their inter-modal correspondence simultaneously \cite{karpathy2014deep}. 
To the best of our knowledge, most methods employ linear mapping functions to translate the image and text feature vectors into a shared embedding space respectively. Although these linear mapping functions are easy to construct, they might not be capable of faithfully reflecting more sophisticated cross-modal correspondence  \cite{jiang2015deep}.
Additionally, given an image query, previous methods suppose that all of the texts in the rank list are of equal importance, and thus either all ranking texts are utilized simultaneously, or a small number of ranking texts are selected randomly at each iteration to train the embedding space. 
Indeed, the texts ranked higher are more accurate, and thus, should be more important than those ranked lower \cite{jiang2014easy}. 
As a result, it is significantly necessary to develop more sophisticated mapping functions and discriminate the contributions of each ranking in a theoretically sound manner.

\begin{figure*}[t]
	\centering
	\includegraphics[scale=.40]{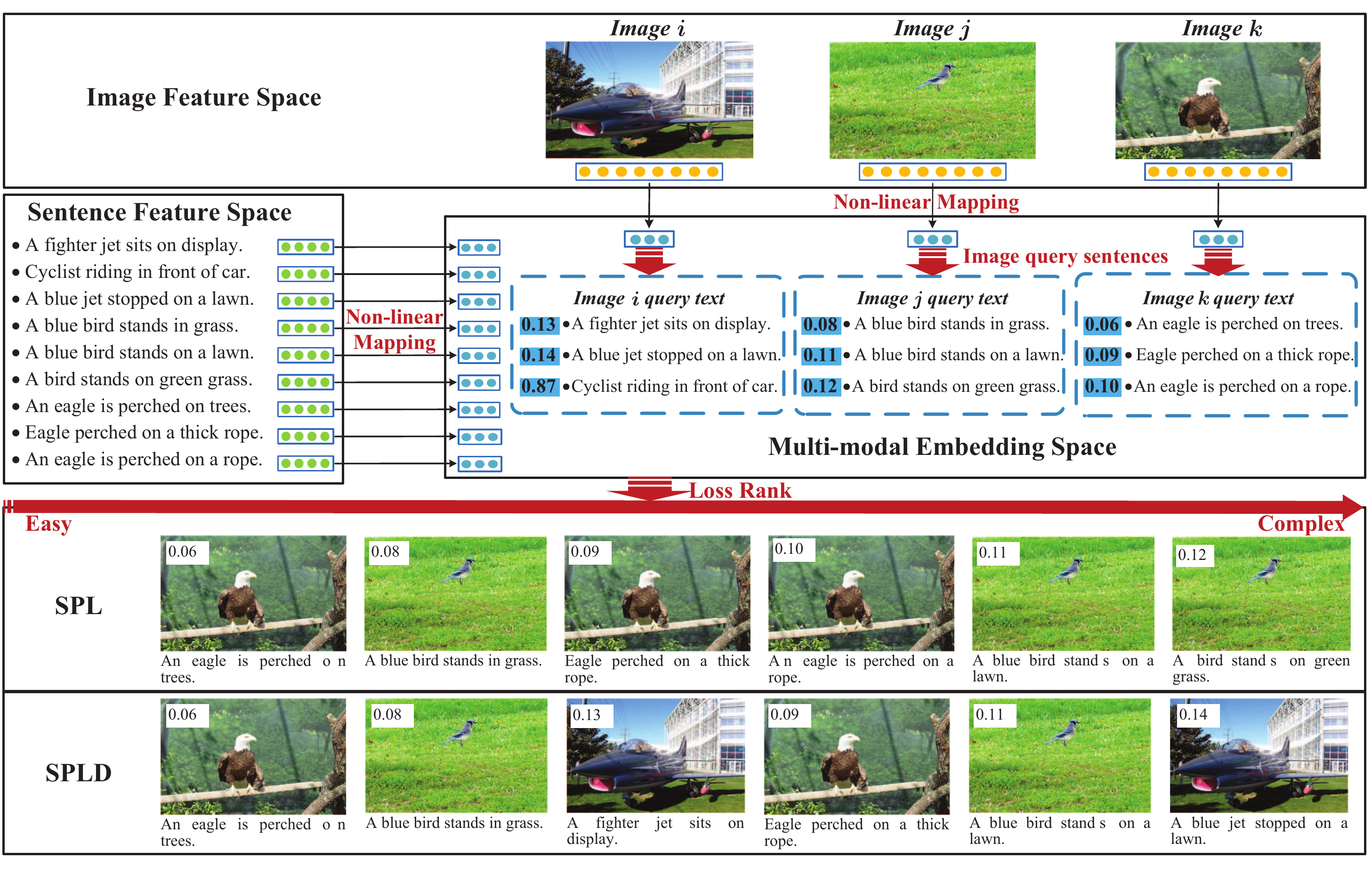}
	\caption{The framework of the proposed simple to complex cross-modal learning to rank.\label{fig:splcmlr}}
\end{figure*}

To this end, we incorporate a self-paced learning with diversity (SPLD) theory \cite{bengio2009curriculum} into CMLR to train an optimal embedding space based on non-linear mapping functions. In such a way, the model is learned gradually from easy rankings with respect to diverse queries to more complex ones.
For a better understanding, we take image-query-sentence as an example to illustrate the proposed framework, as shown in \Cref{fig:splcmlr}. Through non-linear mapping, we translate images and sentences lying in heterogeneous feature spaces into a shared embedding space to facilitate the similarity measurement between image and sentence. 
Given each image query, the retrieved sentences are ordered according to their ranking loss, as specified by the numbers in \Cref{fig:splcmlr}.  
It is reasonable to believe that the sentences ranked higher, \ie, with a smaller loss, are usually more accurate and important.
These ranking sentences with the corresponding image query are referred to as easiness in this paper. To learn an optimal embedding space, we follow the self-paced learning (SPL) and select ranking sentences together with the corresponding image queries from easy to more complex (See the row of SPL in \Cref{fig:splcmlr} for example).
However, SPL only considers about the easiness, not about the diversity of the selected ranking sentences with respect to different image queries.
Indeed, studies have suggested that diversity is an important aspect of learning because performance is enhanced significantly through samples that are dissimilar from what has already been learned \cite{jiang2014self,zhao2015self}. 
Ignoring the diversity may lead to over-fitting to a subset of easy rankings by some specific queries.
This is significant since the over-fitting becomes increasingly severe as the rankings by some specific queries are kept adding into training while ignoring the easy rankings by other queries.  
As shown in the row of SPL in \Cref{fig:splcmlr}, all the rankings with respect to the $i$-th image query fails to be selected with SPL. 
For this issue, we further improve the SPL by considering both easiness and diversity, such that the selected easy rankings are scattered across all image queries as much as possible. 
As indicated in the last row of \Cref{fig:splcmlr}, SPL with diversity (SPLD) helps to select from easy ranking sentences with respect to diverse image queries to more complex ones. 

In summary, we describe the contributions of this paper as follows:
(1) From a new perspective, we adaptively assign each ranking with an importance weight and learn a more optimal multi-modal embedding space gradually from easy to more complex rankings with respect to diverse image queries. (2) We employ non-linear mapping functions to learn the multi-modal embedding space, such that more sophisticated cross-modal correspondence can be captured for cross-media retrieval. (3) An efficient alternative algorithm is exploited to solve the proposed challenging problem with a fast convergence in practice. Extensive experimental results over several benchmark datasets demonstrate the effectiveness and superiority of the proposed algorithm. 

The remainder of this paper is organized as follows. We give a brief introduction to the related work on CMLR and SLP in \Cref{sec:related}. 
In \Cref{sec:cross_media}, we firstly introduce a non-linear mapping to characterize the multi-modalities embedding space, and then we associate each ranking by cross-modal query with an importance weight to train the CMLR model in an SPL fashion with diversity. 
We exploit an efficient alternating algorithm in \Cref{sec:opt} to address the proposed challenging optimization problem.
In \Cref{sec:exp}, we conduct extensive experiments over several benchmark data sets to illustrate the effectiveness and superiority of the proposed method. \Cref{sec:conclusion} concludes this work.

\section{Related Work}
\label{sec:related}
In this section, we briefly review the related work on CMLR and self-paced learning theory and applications. 
\subsection{Cross-media Retrieval}


Grangier \etal pioneered to formalize the cross-modal retrieval tasks as a pair-wise ranking problem and maximize the final retrieval performance with a Passive-Aggressive algorithm, namely Passive-Aggressive Model for Image Retrieval (PAMIR). However, since this method verifies the pairwise ranking criterion with mapping from image query space to the document space, its performance may be deteriorated by the skewed multi-modal data.
Consequently, some efforts are devoted to formalize cross-media retrieval as a list-wise ranking loss optimization problem. 
For example, Xu \etal propose to optimize the list-wise ranking loss with a low-rank embedding; 
Wu \etal \cite{wu2015cross} learn the latent joint representation of multi-modal data through a conditional random field.
Inspired by dictionary learning together with sparse coding techniques, multi-modal dictionary learning is also studied by associating each modal data with a dictionary \cite{monaci2007learning,jia2010factorized}. 
Additionally, hashing technique is also employed to solve the problem of CMLR due to its efficiency for large-scale datasets \cite{zhu2013linear,yu2014hashing, wang2016semantic,caodeep}.

Note that the mentioned approaches above commonly use linear mapping functions to translate multi-modal data into the shared space for its simplicity. However, linear mapping function might not be sophisticated enough to reveal the explicit correspondences between different modalities.
For this reason, Feng \etal \cite{feng2014cross} leverage correspondence autoencoder with deep architectures to learn the mid-level presentation of multi-modal data; Jiang \etal \cite{jiang2015deep} assume a deep compositional cross-modal semantic representation is more attractive for CMLR and optimize the pairwise ranking using non-linear mapping. These techniques have shown their effectiveness to learn a more sophisticated embedding space with large scale training collections. However, an expensive computational cost is usually required by these methods due to a large number of parameters. Additionally, the ranking performance is limited when there is not enough training data available for some real world applications.

It is noteworthy that all of the previous methods suppose the rankings of multi-modal data are of equal importance, without distinguishing each ranking's contribution to multi-modal embedding space learning.
In this paper, we pioneer to assign each ranking an appropriate importance weight and use non-linear mapping to learn an optimal multi-modal embedding space in a self-paced manner.

\subsection{SPL and SPLD}

Enlightened by the learning principle underlying the cognitive process of humans and animals, SPL theory is raised lately to learn the model from easy samples to gradually more complex ones \cite{kumar2010self,meng2017theoretical}. 
This idea is indeed an improvement of the curriculum learning which specifies a sequence of gradually added training samples \cite{bengio2009curriculum}.
Intuitively, since the samples are organized from easiness to hardness instead of using all samples simultaneously or randomly sampling, the SPL (curriculum) can effectively avoid poor local minimum and achieve a better generalization \cite{bengio2009curriculum}.
SPL is independent of particular model objectives and has been attracted increasing attention in the field of machine learning and computer vision tasks, such as object localization and tracking \cite{shi2016weakly,xiao2016track}, reranking of multimedia search \cite{jiang2014easy,liang2017webly}, image classification \cite{tang2012self,tudor2016hard,gong2016multi}, matrix factorization \cite{zhao2015self} and cotraining of multi-view tasks \cite{ma2017self}. 
However, traditional SPL theory just considers the easiness while ignoring the diversity of the selected samples. For this issue, Jiang \etal \cite{jiang2014self} enhance SPL with non-convex diversity regularization such that the selected easy samples dissimilar from what has already been learned; Zhang \etal \cite{zhang2015self} introduce a convex diversity regularization and use SPLD for co-saliency detection. 
In this paper, we incorporate the SPLD into CMLR to training the multi-modal embedding space gradually from easy rankings to more complex ones by diverse queries.

\section{Self-paced Cross-media Learning to Rank}
\label{sec:cross_media}
In this section, we first introduce the framework of CMLR with non-linear mapping functions from image and text feature spaces to a shared embedding space. Then a self-paced CMLR model is proposed with diversity regularization to learn a more optimal embedding space in a theoretically sound manner.

\subsection{Problem Formulation}

We associate each image with a natural language description such that the training dataset consists of $n$ image-text pairs, \ie  $\mathcal{D}=\{(\mathbf{x}_i,\mathbf{z}_i):i=1,2,\cdots,n\}$,
where $\mathbf{x}_i\in\mathcal{\bar{X}}\subseteq R^{p}$ represents a $p$-dimensional visual feature vector extracted from the $i$-th image and $\mathbf{z}_i\in\mathcal{\bar{Z}}\subseteq R^{q}$ refers to a $q$-dimensional feature vector extracted from the $i$-th text (sentence).
For a better representation, we collect all sentences and images in $\mathcal{X}=\{\mathbf{x}_1,\mathbf{x}_2,\cdots,\mathbf{x}_n\}$ and $\mathcal{Z}=\{\mathbf{z}_1,\mathbf{z}_2,\cdots,\mathbf{z}_n\}$, respectively.
Note that the order of components in $\mathcal{X}$ and $\mathcal{Z}$ should correspond with each other such that the $i$-th image $\mathbf{x}_i\in\mathcal{X}$ and the $i$-th text $\mathbf{z}_i\in\mathcal{Z}$ come from the same pair in $\mathcal{D}$.

To explore the underlying correspondence between relevant text and image, a shared multimodal embedding space $\mathcal{\bar{E}}\subseteq R^{d}$ is learned in CMLR to
facilitate the similarity measurement between different modalities.
Given an image query $\mathbf{x}\in\mathcal{X}$, we define a non-linear mapping from image feature space into the shared multimodal embedding space via $h:\mathcal{\bar{X}}\rightarrow \mathcal{\bar{E}}$,
\begin{align}
\label{h}
h(\mathbf{x})=\sigma(W_1\mathbf{x}+\mathbf{b}_1),
\end{align}
where the non-linear mapping $\sigma(\cdot)$ is specified as a Sigmoid function in this paper; $W_1$ is a $d\times p$ transformation matrix and $\mathbf{b}_1\in\mathbb{R}^{d}$ is a bias vector.
Similarly, we map each text feature into the shared embedding space by non-linear mapping $g:\mathcal{\bar{Z}}\rightarrow \mathcal{\bar{E}}$,
\begin{align}
\label{g}
g(\mathbf{z})=\sigma(W_2\mathbf{z}+\mathbf{b}_2),
\end{align}
where $W_2$ is a $d\times q$ transformation matrix and $\mathbf{b}_2\in\mathbf{R}^{d}$ is a bias vector.

Through non-linear mapping $h$ and $g$, the similarity measurement (relevance score) $S(\mathbf{x},\mathbf{z})$ between image query $\mathbf{x}$ and the retrieved text $\mathbf{z}$ can be obtained via computing the cosine similarity in the shared embedding space, i.e.,
\begin{align}
S(\mathbf{x},\mathbf{z})=h(\mathbf{x})^\top g(\mathbf{z}).
\end{align}
In this case, the underlying correspondence between image and text lies in the embedding parameters $W_1,\mathbf{b}_1$ and $W_2,\mathbf{b}_2$. We add one dimension values $1$ in each input feature $\mathbf{x}\in\mathcal{X}$ and $\mathbf{z}\in\mathcal{Z}$ and view the biases $\mathbf{b}_1$ and $\mathbf{b}_2$ as an extra column of the corresponding transformation matrices $W_1$ and $W_2$, respectively. Without loss of generality, we still denote the feature vectors of image and text by $\mathbf{x}$ and $\mathbf{z}$ for a better representation.

To verify the effectiveness of the learned similarity measurement based on embedding parameters, we follow the intuitive strategies used in \cite{jiang2015deep} and assume the aligned text $\mathbf{z}_k$ should ranks higher than the other text $\mathbf{z}_j\in\mathcal{Z}\ (j\neq k)$ given an image query $\mathbf{x}^k\in\mathcal{X}$, i.e.,
\begin{align}
S(\mathbf{x}_k,\mathbf{z}_k)\geq S(\mathbf{x}_k,\mathbf{z}_j)\ \ (\forall j\neq k).
\end{align}
In such a way, we associate each image query $\mathbf{x}_k$ a tetrad set
$\mathcal{T}^k_{\mathbf{x}}=\{(\mathbf{x}_k,\mathbf{z}_k,\mathbf{z}_j,y_{kj}):j=1,2,\cdots,k-1,k+1,\cdots,n\}$,
where $y_{kj}$ is assigned on the basis of similarity measurement $S$ by
\begin{align}
y_{kj}=\left\{
\begin{array}{cl}
1, & \hbox{ $S(\mathbf{x}_k,\mathbf{z}_k)\geq S(\mathbf{x}_k,\mathbf{z}_j)$;} \\
-1, & \hbox{ otherwise.}
\end{array}
\right.
\end{align}
for any $ j\neq k$. As a result, the following inequality should be fulfilled for each tetrad $(\mathbf{x}_k,\mathbf{z}_k,\mathbf{z}_j,y_{kj})\in\mathcal{T}^k_{\mathbf{x}}$,
\begin{align}
\vspace{.5 em}
\label{ys}
y_{kj}\left[S(\mathbf{x}_k,\mathbf{z}_k)-S(\mathbf{x}_k,\mathbf{z}_j)\right]\geq 0.
\vspace{.5 em}
\end{align}
For each ranking text by the $k$-th image query $\mathbf{x}_k$, we define the incurred ranking loss function as
\begin{align}
\vspace{.5 em}
\mathcal{L}(\mathcal{T}^k_{\mathbf{x}}; W)
=\sum_{j\neq k}
\ l(\mathbf{x}_k,\mathbf{z}_k,\mathbf{z}_j,y_{kj}; W),
\vspace{.5 em}
\end{align}
where $W=\{W_1,\mathbf{b}_1,W_2,\mathbf{b}_2\}$ collects the embedding parameters used in functions (\ref{h}) and (\ref{g}); $l(\mathbf{x}_k,\mathbf{z}_k,\mathbf{z}_j,y_{kj}; W)$ is usually given as a hinge loss by
\begin{align}
\label{l}
l(\mathbf{x}_k,\mathbf{z}_k,\mathbf{z}_j,Y_{kj};W)
=\max(0,y_{kj}[S(\mathbf{x}_k,\mathbf{z}_j)-S(\mathbf{x}_k,\mathbf{z}_k)]+\Delta)
\end{align}
with margin $\Delta\geq0$. This objective loss encourages aligned image-text pairs in $\mathcal{D}$ to have a higher score than misaligned pairs by a margin. 

\subsection{Self-paced CMLR with Diversity Regularization}

In this part, we incorporate the SPLD into CMLR framework to enhance the embedding space learning. This idea tends to distinguish faithful tetrads $(\mathbf{x}_k,\mathbf{z}_k,\mathbf{z}_j,y_{kj})$ from easy (high-confidence) ones, and then gradually transfer the learning knowledge to recognize more complex ones.
To this end, we collect the entire tetrads over all image queries into set $\mathcal{T}_{\mathcal{X}}$,
\begin{align}
\mathcal{T}_{\mathcal{X}}=\{(\mathbf{x}_k,\mathbf{z}_k,\mathbf{z}_j,y_{kj})\in\mathcal{T}_{\mathbf{x}}^k:j\neq k;k=1,2,\cdots,n\}
\end{align}
and assign each tetrad $(\mathbf{x}_k,\mathbf{z}_k,\mathbf{z}_j,y_{kj})\in\mathcal{T_{\mathcal{X}}}$ a weight $v^k_{j}$ to reflect the importance of ranking text $\mathbf{z}_j$ by image query $\mathbf{x}_k$. 
Specifically, the importance weight vector  $\mathbf{v}\in\mathbb{R}^{n(n-1)}$ over set $\mathcal{T}_{\mathcal{X}}$ is defined as
\begin{align*}
\mathbf{v}^\top=[\underbrace{v^1_2,v^1_3,\cdots,v^1_n}_{\mathbf{v}^1\in\mathbb{R}^{n-1}},
\underbrace{v^2_1,v^2_3,\cdots,v^2_n}_{\mathbf{v}^2\in\mathbb{R}^{n-1}},
\cdots,
\underbrace{v^n_1,v^n_2,\cdots,v^n_{n-1}}_{\mathbf{v}^n\in\mathbb{R}^{n-1}}].
\end{align*}
In particular, given image query $\mathbf{x}_k$, the loss incurred by ranking text $\mathbf{z}_j$ has no effect on embedded space learning if $v^k_{j}=0$, i.e., the tetrad $(\mathbf{x}_k,\mathbf{z}_k,\mathbf{z}_j,y_{kj})$ will not be evolved in the procedure of training.

With the importance weight vector $\mathbf{v}$, the idea of self-paced CMLR with diversity regularization is formalized as solving the following optimization problem:
\begin{align}
\label{self_obj}
&\min_{W,\mathbf{v}} \ \frac{1}{2}\|W\|^2+\sum_{k}\sum_{j\neq k}
v^k_{j}\ l(\mathbf{x}_k,\mathbf{z}_k,\mathbf{z}_j,y_{kj}; W)
+\phi(\mathbf{v};\lambda,\gamma)\\
&\ \ \emph{s.t.}\ \ \ v^k_{j}\in[0,1]\ \ \ \ (\forall k,\ j\neq k)\notag
\end{align}
where the self-pace capability with diversity is achieved through regularization function
\begin{align}
\label{phi}
\phi(\mathbf{v};\lambda,\gamma)
&=-\lambda\|\mathbf{v}\|_1-\gamma\|\mathbf{v}\|^*_{2,1}
\end{align}
where $\lambda$ and $\gamma$ are the regularizer penalty parameters which are imposed on the negative $l_1$-norm term (easiness term) and the $l_{2,1}$-norm-like term $\|\mathbf{v}\|^*_{2,1}$ (diversity term), respectively.
Specifically, the easiness term is defined as
\begin{align}
-\lambda\|\mathbf{v}\|_1=-\lambda\sum_{k}\sum_{j\neq k} v^k_{j}.
\end{align}
It favors selecting from easy tetrads in $\mathcal{T}_{\mathcal{X}}$ to more complex ones.
Without considering the diversity term, i.e., $\gamma=0$, the importance weight $v^k_j\in[0,1]$ is updated for each tetrad $(\mathbf{x}_k,\mathbf{z}_k,\mathbf{z}_j,y_{kj})$ with fixed embedding parameters $W$, according to
\begin{align}
\label{vkj}
v^k_j=\left\{
\begin{array}{ll}
1, & \hbox{$l(\mathbf{x}_k,\mathbf{z}_k,\mathbf{z}_j,y_{kj}; W)\leq\lambda$;} \\
0, & \hbox{otherwise.}
\end{array}
\right.
\end{align}
As a result, the tetrad with smaller loss is taken as an easy one and therefore should be learned preferentially by setting $v^k_j=1$ and vice versa. According to this strategy, the parameter set $W$ is updated iteratively only on the selected preferable tetrads with importance weight $v^k_j=1$. As the increase of $\lambda$, more tetrads with larger loss will be gradually involved to learn a more ``nature'' model \cite{jiang2014easy}. Thus, by incorporating the estimation of importance weight vector $\mathbf{v}$ with negative $l_1$-norm regularization into the procedure of CMLR, we indeed achieve to learn the embedding of cross-media data in a self-paced fashion.

However, the easiness regularization might lead to some importance weight vector, for instance $\mathbf{v}^{k_0}$ with respect to the $k_0$-th textual query, becomes zero vector, i.e., $v^{k_0}_j=0$ for $j=1,2,\cdots,k_0-1,\cdots,k_0+1,\cdots,n$. It leads that all of the tetrads in set $\mathcal{T}_{\mathbf{x}}^{k_0}$ for the $k_0$-th image query are never selected to update embedding parameter $W$ in the next iteration.
This is because the updating of importance weight vector $\mathbf{v}$ according to (\ref{vkj}) does not consider the diversity of the selected tetrads, where the diversity is an important aspect of learning because performance is usually enhanced significantly through samples that are dissimilar from what has already been learned \cite{jiang2014self,zhao2015self}.
For this issue, it is necessary to impose the diversity regularization on the importance weights vector, such that the selected tetrads are scattered over different image queries. 
In this paper, we following the strategy used in \cite{zhang2015self} and define the diversity regularization as a $l_{2,1}$-norm-like term $\|\mathbf{v}\|^*_{2,1}$, 
\begin{align}
\label{21normlike}
\|\mathbf{v}\|^*_{2,1}=\sum_{k}\sqrt{\sum_{j\neq k}v^{k}_j}.
\end{align}
Intuitively, the diversity regularization evolved into the objective function (\ref{self_obj}) aims to prevent the non-zero importance weights from concentrating in some image queries and ignoring others, i.e., to select dissimilar tetrads from different image queries as much as possible.
Note that the diversity regularization defined in (\ref{21normlike}) makes $\|\mathbf{v}\|^*_{2,1}$ non-convex, while guaranteeing the convexity of its negative. This strategy preserves the previous axiomatic definition for SPL regularization. Moreover, it makes the solution of the optimization problem (\ref{self_obj}) more sufficient and simple. 

In summary, we pioneer to associate each tetrad with an adaptive importance weight and employ self-pace regularization $\phi(\mathbf{v},\lambda,\gamma)$ to guide the learning in a theoretically sound manner. This strategy enhances model's robustness to outliers and improve its generalization capability.

\section{Optimization Procedure}
\label{sec:opt}
In this section, we exploit an alternative optimization algorithm to solve the proposed challenging problems via updating embedding parameters $W$ and importance vector $\mathbf{v}$ iteratively with the other one fixed.

\subsection{Optimize $W$}

In this step, we seeks to estimate the embedding parameters, i.e., $W_1,\mathbf{b}_1$ and $W_2,\mathbf{b}_2$. With fixed importance weight vector $\mathbf{v}$, the optimization problem (\ref{self_obj}) degenerates to the following form: 
\begin{align}
\label{w}
\min_{W} \ \frac{1}{2}\|W\|^2+\sum_{k}\sum_{j\neq k}
v^k_{j}\ l(\mathbf{x}_k,\mathbf{z}_k,\mathbf{z}_j,y_{kj}; W).
\end{align}
For this optimization problem, we use the gradient descent method to update embedding parameters $W$ at each iteration. Let the objective in optimization problem (\ref{w}) be $f(W;\mathbf{v})$. 
The derivatives of $f(W;\mathbf{v})$ with respect to parameter $W_1$, denoted by $\nabla f_{W_1}$, can be computed as 
\begin{align}
\label{w1}
\nabla f_{W_1}
&=\frac{\partial f(W;\mathbf{v})}{\partial W_1}\notag\\
&= W_1 + \sum_{k}\sum_{j\neq k}
v_{kj}\frac{\partial l(\mathbf{x}_k,\mathbf{z}_k,\mathbf{z}_j,y_{kj}; W)}{\partial W_1}
\end{align}
where according to the definition of loss function in (\ref{l}), we calculate its gradient with respect to parameter $W_1$ as follows
\begin{align}
\vspace{.5 em}
\label{lw1}
\frac{\partial l(\mathbf{x}_k,\mathbf{z}_k,\mathbf{z}_j,y_{kj}; W)}{\partial W_1}=
y_{kj}(\frac{\partial S(\mathbf{x}_k,\mathbf{z}_i)}{\partial W_1}-\frac{\partial S(\mathbf{x}_k,\mathbf{z}_j)}{\partial W_1}).
\vspace{.5 em}
\end{align}
From the gradient above, we observe that the cost function value is back propagated into the gradient of similarity measurement $S(\mathbf{x},\mathbf{z})$ with respect to parameter $W_1$, \ie
\begin{align}
\vspace{.5 em}
\label{sw1}
\frac{\partial S(\mathbf{x},\mathbf{z})}{\partial W_1}
&=\frac{\partial {\left( h(\mathbf{x})^\top g(\mathbf{z}) \right)}}{\partial W_1}\notag\\
&=\frac{\partial {\left({f(W_1 \mathbf{x}+\mathbf{b}_1)}^\top g(W_2 \mathbf{z}+\mathbf{b}_2)\right)}}{\partial W_1}\notag\\
&=\left( g(W_2 \mathbf{z}+\mathbf{b}_2)\odot f'(W_1 \mathbf{x}+\mathbf{b}_1)\right) {\mathbf{x}}^\top
\vspace{.5 em}
\end{align}
where $\odot$ represents the element-wise multiplication and $f'(\cdot)$ refers to the derivative of $f(\cdot)$ with respect to its input. Based on the three equations above, we achieve the gradient of $\Omega(W;\mathbf{v})$ with respect to parameter $W_1$. According to similar derivations, we can also obtain the gradient of $f(W;\mathbf{v})$ with respect to parameter $W_2$, denoted by $\nabla f_{W_2}$. In summary, the embedding parameters $W_1$ and $W_2$ can be updated as
\begin{align}
W_1&\leftarrow W_1+a\nabla f_{W_1}\label{upw1}\\
W_2&\leftarrow W_2+b\nabla f_{W_2}\label{upw2}
\end{align} 
where $a$ and $b$ are the step size which can be found by linear search.

\subsection{Optimize $\mathbf{v}$}
After updating the embedding parameters, we renew the weights $v_{kj}\ (j\neq k)$ to reflect the adaptive importance of tetrad $(\mathbf{x}_k,\mathbf{z}_k,\mathbf{z}_j,y_{kj})$. Following the algorithm proposed in \cite{zhang2015self}, when $W$ is fixed, $\mathbf{v}$ can be updated by solving optimization problem
\begin{align}
\label{v}
&\min_{\mathbf{v}} \ \sum_{k}\sum_{j\neq k}
v^k_{j}\ l(\mathbf{x}_k,\mathbf{z}_k,\mathbf{z}_j,y_{kj}; W)
+\phi(\mathbf{v};\lambda,\gamma)\\
&\ \ \emph{s.t.}\ \ \ v^k_{j}\in[0,1]\ \ \ \ (\forall k,\ j\neq k)\notag
\end{align}
where $\phi(\mathbf{v};\lambda,\gamma)
=-\lambda\sum_{k}\sum_{j\neq k} v^k_{j}-\gamma\sum_{k}\sqrt{\sum_{j\neq k}v^{k}_j}$.
Let $l_j^k=l(\mathbf{x}_i,\mathbf{z}_k,\mathbf{z}_j,y_{kj}^k;$ $ W)$. Since the objective function (\ref{v}) is independent between
different $k$, we can estimate the importance weight  $\mathbf{v}^k=[v^k_1,\cdots,v^k_{k-1}, $ $v^k_{k+1},\cdots,v^k_n]$ individually via solving the following optimization problems
\begin{align}
\label{vk}
&\min_{\mathbf{v}^k} \ \psi(\mathbf{v}^k)=\sum_{j\neq k}
v^k_{j}\ l_j^{k}
-\lambda\|\mathbf{v}^k\|_1-\gamma\sqrt{\sum_{j\neq k}v^{k}_j}\\
&\ \ \emph{s.t.}\ \ \ v^k_{j}\in[0,1]\ \ \ \ (\ j\neq k)\notag
\end{align}
for each $k=1,2,\cdots,n$. In terms of Lagrangian parameters
\begin{align}
\alpha^k&=[\alpha^k_1,\cdots,\alpha^k_{k-1},\alpha^k_{k+1},\cdots,\alpha_n]^\top\in\mathbb{R}^{n-1};\\
\beta^k&=[\beta^k_1,\cdots,\beta^k_{k-1},\beta^k_{k+1},\cdots,\beta_n]^\top\in\mathbb{R}^{n-1},
\end{align} 

\begin{algorithm}[t]
	\caption{Algorithm of Optimizing Importance Weight $\mathbf{v}$. \label{alg:optv}}
	\begin{algorithmic}[1]
		\REQUIRE Tetrads set $\mathcal{T}_{\mathcal{X}}=\{(\mathbf{x}_k,\mathbf{z}_k,\mathbf{z}_j,y_{kj})\in\mathcal{T}_{\mathbf{x}}^k:j\neq k;k=1,2,\cdots,n\}$; Current embedding parameters $W$; two trade-off parameters $\lambda$ and $\gamma$.
		\ENSURE The global optimum $\mathbf{v}^*$ of optimization problem (\ref{v}).
		\FOR{$k=1,2,\cdots,n$}
		\STATE Sort the Tetrads in ascending order according to their loss values as
		$l^k_{j_1}\leq l^k_{j_2}\leq \cdots\leq l^k_{j_{n-1}};$		
		\STATE  Compute $U=\{u:l^k_{j_u}<\lambda+\frac{\gamma}{2\sqrt{j_u}}\}$;
		\STATE \textbf{if} $U\neq\emptyset$\ \ \textbf{then}\ \  $v^k_{j_u}=1,\forall u\in U$;
		\STATE  Find $u^\prime=\arg\min_{1,2,\cdots,n-1}\{u:\ v^k_{j_u}=v^k_{j_{u+1}}=\cdots=v^k_{j_{u+m-1}},
		l^k_{j_u}\geq\lambda+\frac{\gamma}{2\sqrt{j_u}}\}$ and set
		$v^k_{j_{u^\prime+m}}=v^k_{j_{u^\prime+m+1}}=\cdots=v^k_{j_{n-1}}=0$,
		\begin{align*}
		&v^k_{j_{u^\prime}}
		=\cdots=v^k_{j_{u^\prime+m-1}}
		=\frac{\gamma/2(l^k_{j_{u^\prime}}-\lambda)^2-j_{u^\prime}+1}{m};
		\end{align*} 
		\ENDFOR
	\end{algorithmic}
\end{algorithm}
\begin{algorithm}[t]
	\caption{Algorithm for Optimization Probelem (\ref{self_obj})\label{alg:spl}}
	\begin{algorithmic}[1]  
		\REQUIRE Tetrads set $\mathcal{T}_{\mathcal{X}}=\{(\mathbf{x}_k,\mathbf{z}_k,\mathbf{z}_j,y_{kj})\in\mathcal{T}_{\mathbf{x}}^k:j\neq k;k=1,2,\cdots,n\}$; two trade-off parameters $\lambda$ and $\gamma$.
		\ENSURE parameters $W$.
		\REPEAT
		\STATE Updating $W$ according to (\ref{upw1}) and (\ref{upw2})
		\STATE Updating $\mathbf{v}$ using \Cref{alg:optv}.
		\UNTIL {convergence}	
	\end{algorithmic}
\end{algorithm}
\noindent the Lagrangian function of $\psi(\mathbf{v}^k)$ is formulated as
\begin{align}
L(\mathbf{v}^k,\alpha^k,\beta^k)=\psi(\mathbf{v}^k)-\sum_{j}\alpha^k_jv_j^k-\sum_{j\neq k}\beta^k_j(1-v^k_j).
\end{align}
Consequently, we arrive at the corresponding KKT conditions \cite{boyd2004convex} as: 
\begin{align}
\frac{\partial L}{\partial \mathbf{v}_j^k}
&=l^k_j-\lambda-\frac{\gamma}{2\sqrt{\sum_{j\neq k}v^{k}_j}}-\alpha_j^k+\beta^k_j=0\label{kkt1}\\
\alpha_j^kv^k_{j}&=0;\label{kkt2}\\
\beta^k_j(1-v^k_j)&=0;\label{kkt3}\\
\alpha_j^k\geq 0;\ &\ \beta^k_j\geq 0. \label{kkt4}
\end{align}
Thanks to the convexity of objective function $\psi(\cdot)$, we get a global optimum $\mathbf{v}^*$ satisfying these KKT conditions (\ref{kkt1})-(\ref{kkt4}). In summary, we describe the algorithm of optimizing $\mathbf{v}$ in \Cref{alg:optv}. The overall alternative optimization algorithm for the proposed self-paced CMLR with diversity regularization is summarized in \Cref{alg:spl}. The proposed algorithm is efficient due to the following analysis.
Let $z$ be the average number of tetrads selected by the self-paced function. In each iteration, the complexity mainly lies in Step 2 and 3 of \Cref{alg:spl}. In Step 2, the main computational overhead comes from obtaining similarity score for the selected tetrads with complexity $O(z^2)$. In Step 3, the main computational cost of updating $\mathbf{v}$ lies in the calculation of loss of each tetrad with complexity $O(n^2)$. 

\section{Experiments}
\label{sec:exp}

To illustrate the effectiveness and superiority of the proposed \textbf{S}imple to \textbf{C}omplex \textbf{C}ross-\textbf{M}odel learning to rank framework (\textbf{SCCM}), we perform extensive experiments over some benchmark datasets and demonstrate the efficiency of the diversity regularization used in self-paced learning.
\subsection{Dataset Description and Experimantal Setup}
\begin{figure}[t]
	\centering
	\subfloat[Pairs of image-text examples in Pascal'07 dataset]{\includegraphics[scale=.36]{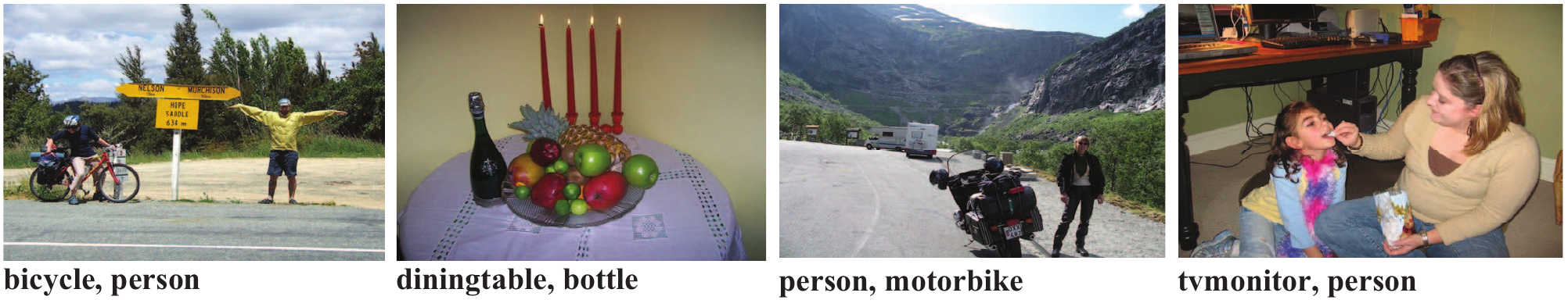}}  \\
	\subfloat[Pairs of image-textexamples in NUS-WIDE dataset]{\includegraphics[scale=.36]{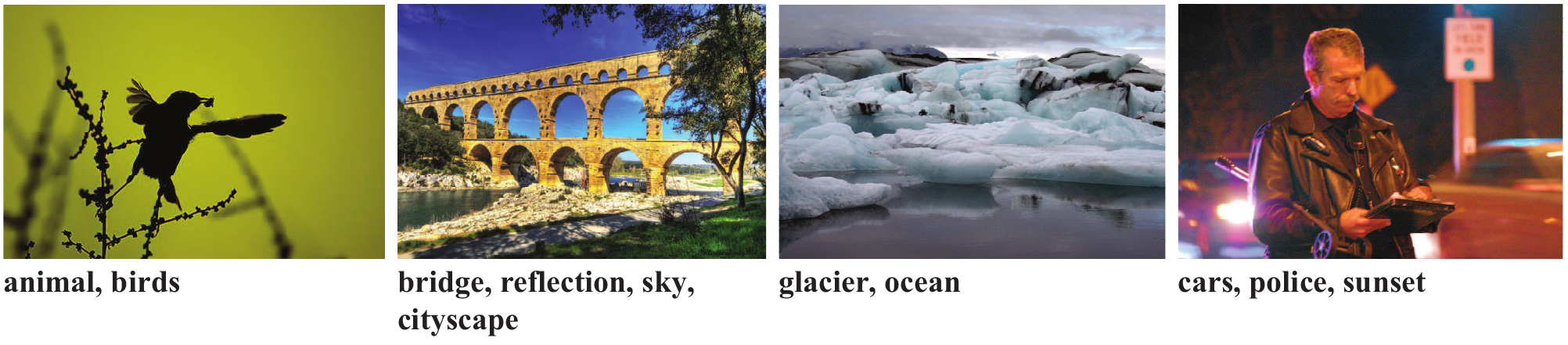}} \\
	\subfloat[Pairs of image-text examples in Wiki dataset]{\includegraphics[scale=.36]{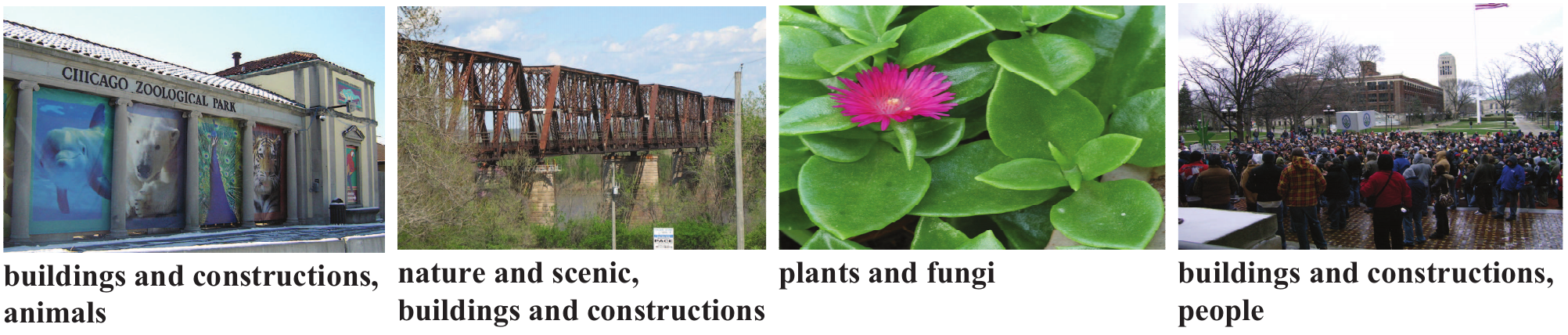}}
	\caption{The pairs of image-text examples in Pascal'07, NUS-WIDE and Wiki datasets, respectively.}
	\label{fig:example}
	\vspace{-1em}
\end{figure}

The details of the datasets are introduced as follows, together with the feature representation of the texts for each dataset. 
Some example pairs of image-text are shown in \Cref{fig:example} for a better understanding.

\begin{enumerate}[$\bullet$,leftmargin=*,topsep=0pt,noitemsep]
	\item[--] \textbf{Pascal'07:} The Pascal'07 dataset is a widely used benchmark dataset in category recognition and multi-modal classification. It consists of 10,000 images from 20 different categories. 804 corresponding tags are downloaded from Flickr for each image in the dataset and are represented as an 804-dimensional feature vector, each of whose dimension indicates if a tag appears. There are 9,587 images in the dataset with the user tags available, which are the image-tag pairs we use in the experiment.
	
	\item[--] \textbf{NUS-WIDE:} This dataset contains 269,648 images with 1,000 associated tags from Flickr. Each image with the corresponding tags has several of the 81 concepts as the ground-truth. We represent the corresponding tags of each image as a 1000-dimensional vector, each dimension of which is a binary indicator to indicate whether a tag appears or not. 
	
	\item[--] \textbf{Wiki image-text:} The dataset contains 2,886 articles with the corresponding image in each of the articles.
	All of the Wikipedia articles are categorized as one of the ten semantic classes. We extract a feature vector from each article with the bag of words model (BoW), resulting a 1000-dimensional representation.
\end{enumerate}

For a fair comparison, we use the same experimental setting in \cite{RasiwasiaPCDLLV10}. Specifically, for Pscal'07 and NUS-WIDE datasets, 2000 images and the associated tags are randomly selected as the training set. 1000 images and the corresponding tags are selected as a validation set used for parameter tuning. The rest are used for testing. For the Wiki dataset, 1200 images and the corresponding text documents are randomly selected as the training set. 500 images and the corresponding text documents are selected as a validation set used for the parameter tuning. Note that the text consisting of multiple tags is used as a query in the experiments. 

\subsection{Feature Extraction}

Previous researches have domonstrated that pre-trained model with ImageNet dataset can boost the performance of other important tasks. Different from the leading approaches, who all pre-train CNN models from the 1,000 classes defined in the ImageNet Large Scale Visual Recognition Challenge, we leverage the complete ImageNet hierarchy for pre-training deep networks following \cite{MettesKS16}. The key insight in \cite{MettesKS16} is that by utilizing the graph structure of ImageNet to combine and merge classes into balanced and reorganized hierarchies, a significant improvement on the visual recognition task can be achieved. To deal with the problems of over-specific classes and imbalanced classes, we adopt a bottom-up and top-down approach for reorganization of the ImageNet hierarchy. After the training data has been re-organized, we pre-train a CNN model using the same architecture as GooLeNets \cite{SzegedyLJSRAEVR15}. The Caffe toolkit \cite{jia2014caffe} is used in our experiment. After pre-training, we extract features at the pool5 layer, with a 1,024-dimensional frame representation. We normalize the representation by $\ell_2$-normalization.

\subsection{Competitors}

To evaluate the effectiveness of the proposed method \textbf{SCCM}, we compare with the following alternatives. We choose the best parameters using 5-fold cross-validation. All the parameters are tuned in the range of $\{10^{-3},10^{-2},10^{-1},10^{0},10^{1}, $ $10^{2},10^{3}\}$.

\begin{enumerate}[$\bullet$,leftmargin=*,topsep=0pt,noitemsep]
	\item[--] \textbf{CCA} \cite{hardoon2004canonical}: CCA maps the pairs of images and text texts into a latent space,  and therefore, the latent representations of the images and text texts are obtained. After the latent representations are individually obtained, CCA performs cross-modal retrieval by measuring the relevance of queries and texts with the cosine similarity regarding their individually latent representation.
	
	\item[--] \textbf{C-CRF} \cite{QinLZWL08}: C-CRF first maps the pairs of images and text texts into a latent space, followed by performing the cross-modal retrieval with C-CRF in a list-wise ranking manner.
	
	\item[--] \textbf{PAMIR} \cite{grangier2008discriminative}: PAMIR utilizes global alignment to learn the latent representations of pairs of images and text texts in a pairwise ranking manner. Image and texts are embedded into a global common space using a linear projection.
	
	\item[--] \textbf{SSI} \cite{bai2010learning}: SSI discriminatively trains a class of nonlinear models to map from the word content in a query-document or document-document to a ranking score in a pairwise ranking manner.
	
	\item[--] \textbf{CMRNN} \cite{Lu2014learning}: CMRNN is built on top of neural networks and learning to rank techniques, which learns high-level feature representation with discriminative power for cross-modal ranking.
	
	\item[--] \textbf{DeepFE} \cite{karpathy2014deep}: DeepFE learns a multi-modal embedding space for fragments of images and sentences and reasons about their latent, inter-modal alignment. It considers the local alignment of images and sentences.
	
	\item[--] \textbf{C$^2$MLR} \cite{jiang2015deep}: C$^2$MLR considers learning a multi-modal embedding from the perspective of optimizing a pairwise ranking problem while enhancing both local alignment and global alignment. C$^2$MLR learn a ranking manner using the local common space and the global common space jointly, where the local common space is computed by local alignment of visual objects and textual words and the global common space is from the global alignment of images and text.
\end{enumerate}

\subsection{Evaluation Metric}

Mean Average Precision (mAP) is used as an evaluation metric, which is one of the most widespread performance evaluations of information retrieval. Given the Average Precision (AP) of all queries, mAP is the mean of all AP values. And the value AP of a query is calculated according to the formula (\ref{ap}).
\begin{align}
\label{ap}
AP(Y,Y')=\frac{1}{R}\sum_{j=1}^RPrec(j)Rel(j)
\end{align}
where $Y$ and $Y'$ denotes the true ranking list and the predicted ranking list namely;
$R$ is the number of retrieved texts to be examined in the ranking list if $R$ is 5, the mAP is represented mAP@5,
and when the value of $R$ is the number of all texts, the mAP is represented mAP@all;
$Prec(.)$ is the percentage of the relevant texts in the top $j$ texts in the predicted ranking;
$Rel(.)$ is the indicator function equaling to 1 if the document at rank $j$ is relevant. mAP is a more suitable measure than other mentioned metrics in this particular task. 
Note that mAP indeed considers both precision and recall of the retrieved results, and thus it is a suitable measure for specific task of cross-media learning to rank.

\subsection{Performance Comparison}

\begin{figure}[t]
	\centering
	\subfloat[Pascal'07 Image-Query-Text]{\includegraphics[scale=0.23]{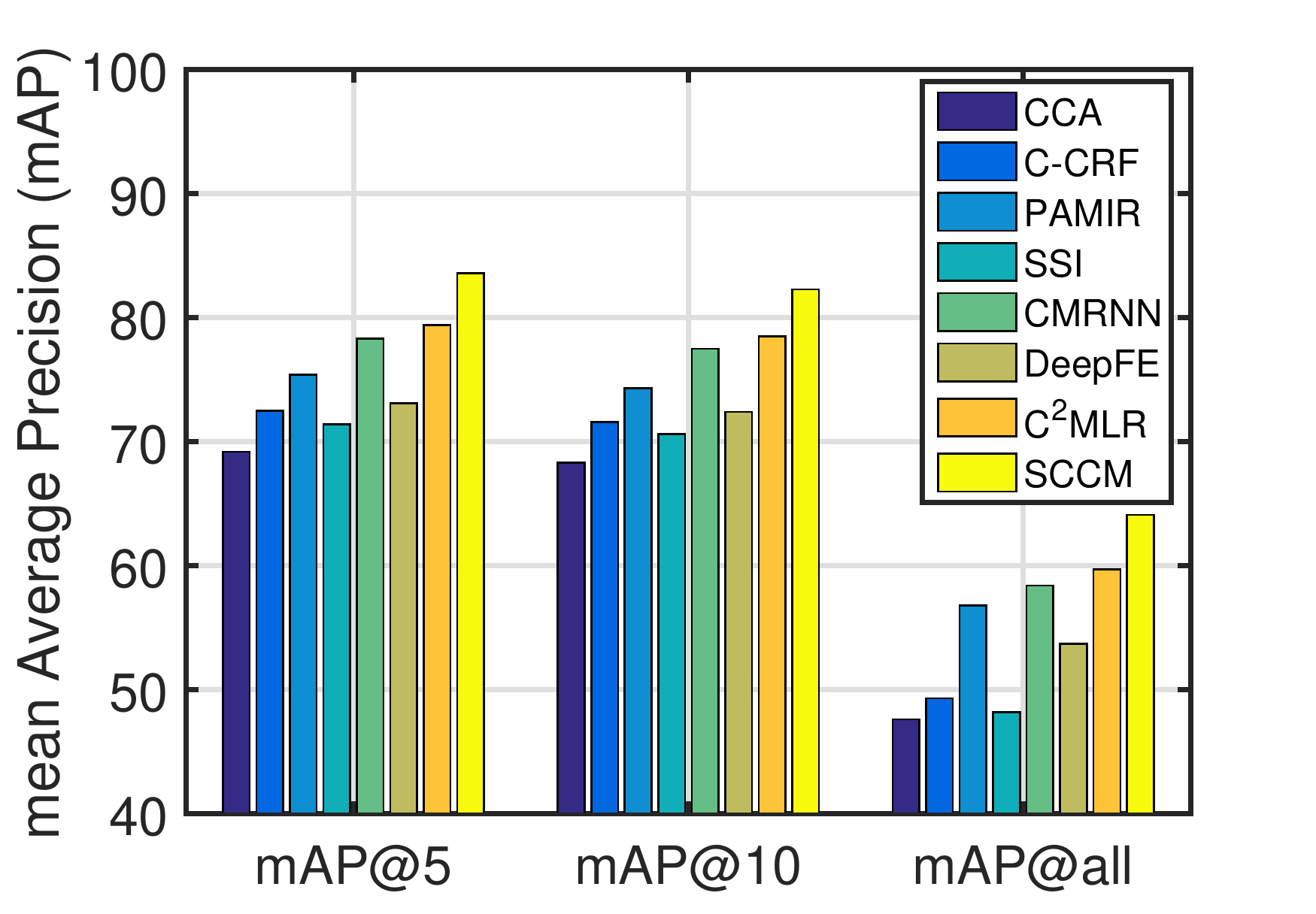} \label{fig:pascal07_image2text}} 
	\subfloat[Pascal'07 Text-Query-Image]{\includegraphics[scale=0.23]{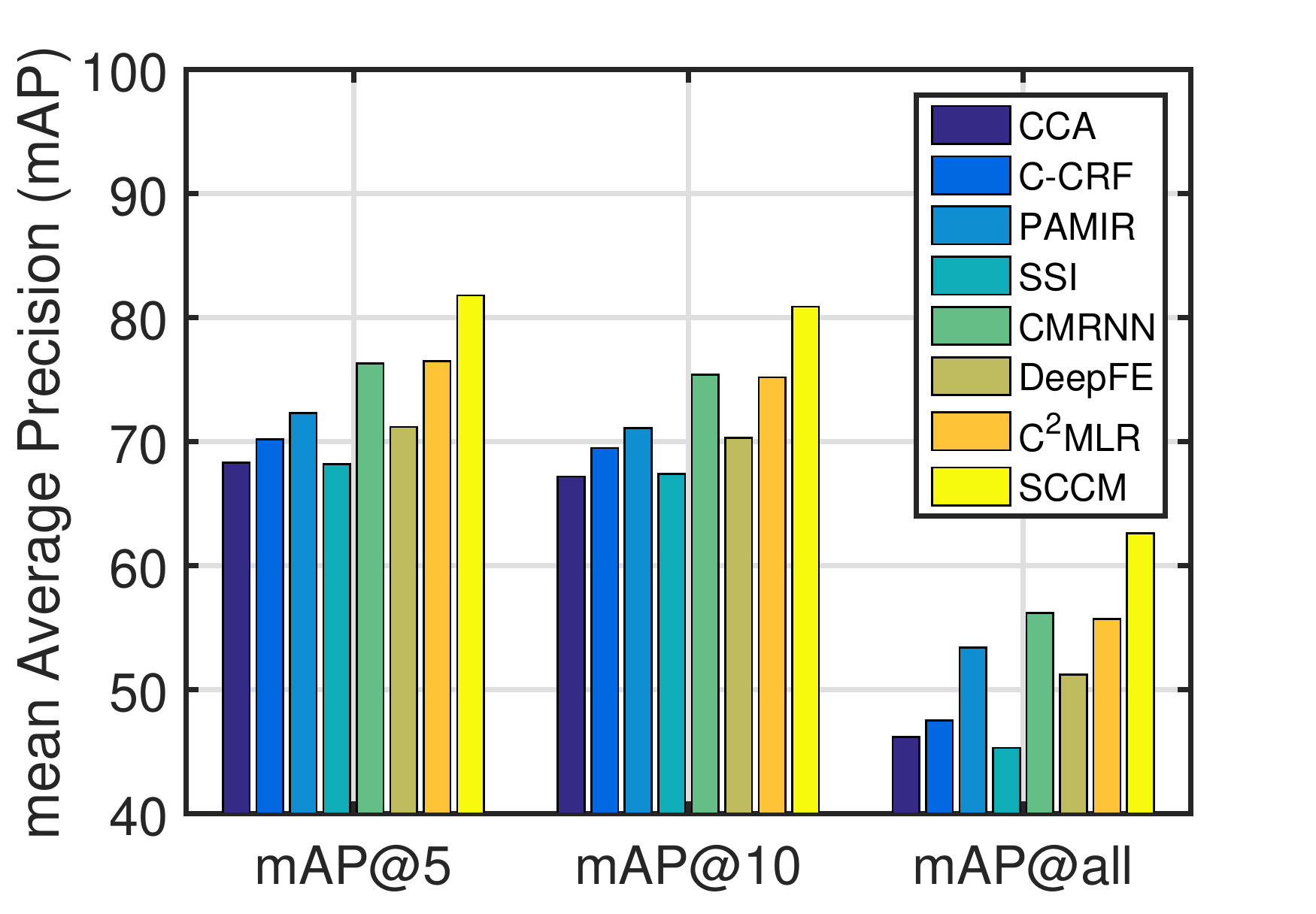}\label{fig:pascal07_text2image}}
	\vspace{-0.5em}
	\caption{The performance comparison in terms of mAP over Pascal'07 dataset.}
	\vspace{-0.5em}
\end{figure}

We report the performance of cross-modal ranking in terms of mAP on Pascal'07 dataset in \Cref{fig:pascal07_image2text} (Image-Query-Text) and \Cref{fig:pascal07_text2image} (Text-Query-Image). Note that whenever possible we have quoted the numbers directly from the references, while if not available we used code from the respective authors to obtain the results ourselves.
From \Cref{fig:pascal07_image2text} and \Cref{fig:pascal07_text2image}, it can be seen that the proposed algorithm outperforms the other alternatives by a large margin. We have the following observations on Pascal'07:

\begin{enumerate}[$\bullet$,leftmargin=*,topsep=0pt,noitemsep]
	\item[--] The performances of all the baseline algorithms are much better than those reported in \cite{jiang2015deep, Lu2014learning}. We attribute this improvement to the adoption of ImageNet Shuffle model for feature extraction. To the best of our knowledge, this is the first work to use ImageNet Shuffule model for learning to rank algorithm.
	
	\item[--] CCA obtains very similar performance in both directions of the retrieval. This is because CCA learns the joint representation from paired multi-modal data in which the pair-correspondence of images and text texts ensures an equal contribution to the learned metric in both modalities.
	
	\item[--] PAMIR performs much better than CCA and C-CRF in both directions, which confirms that learning a good representation for multi-modal data is crucial for cross-modal ranking.
	
	\item[--] We observe that cross-modal with local alignment (\ie, DeepFE) obtains a poor performance on this dataset. This is because many annotated tags of images in Pascal'07 dataset do not explicitly align with visual objects in images.
	
	\item[--] The proposed \textbf{SCCM} outperforms all baseline methods, which confirms the assumption that learning from simple to complex and considering diversity for learning to rank is instrumental. 
\end{enumerate}

\begin{figure}[t]
	\centering
	\subfloat[NUS-WIDE Image-Query-Text]{\includegraphics[scale=0.23]{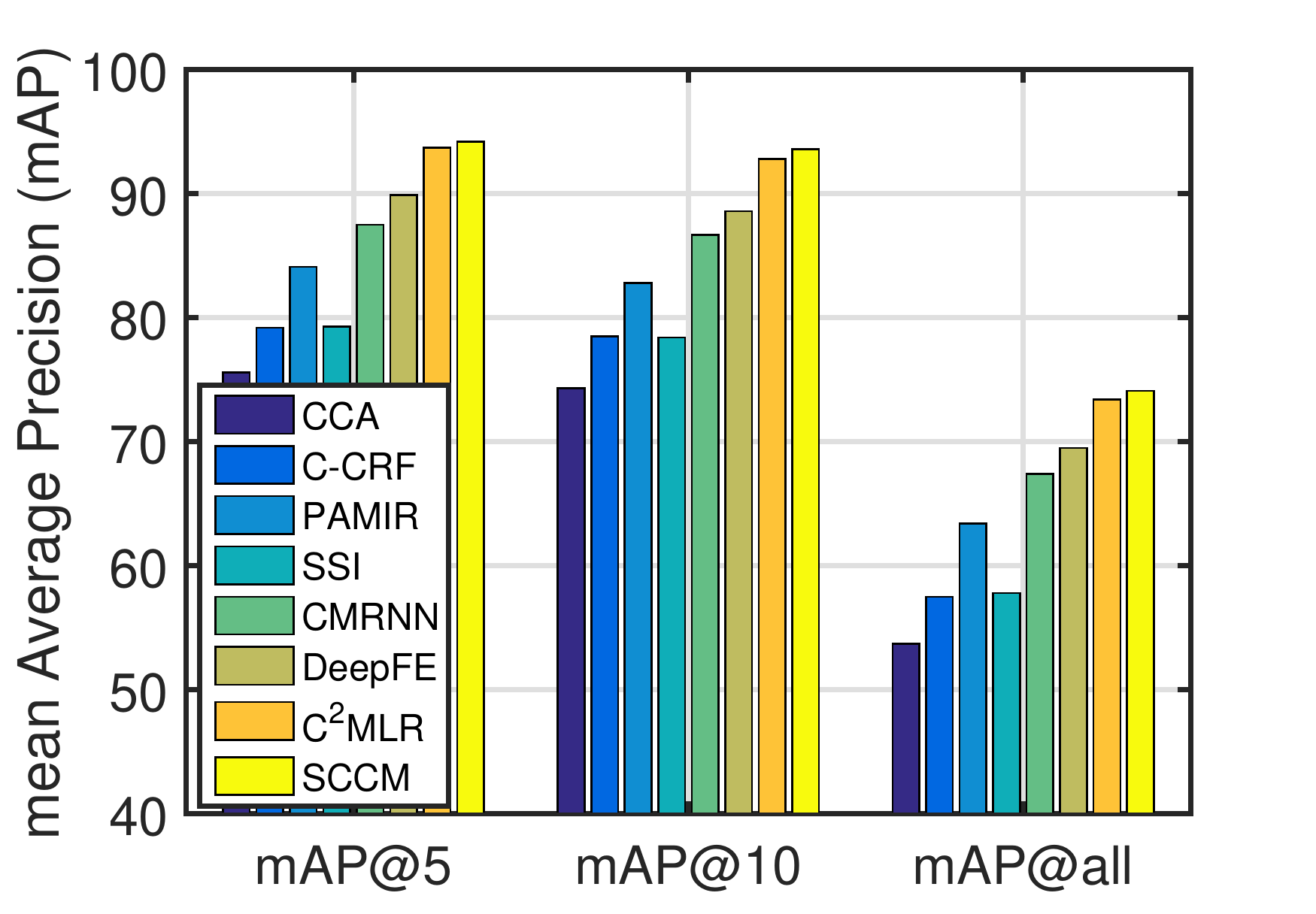} \label{fig:nuswide_image2text}}
	\subfloat[NUS-WIDE Text-Query-Image]{\includegraphics[scale=0.23]{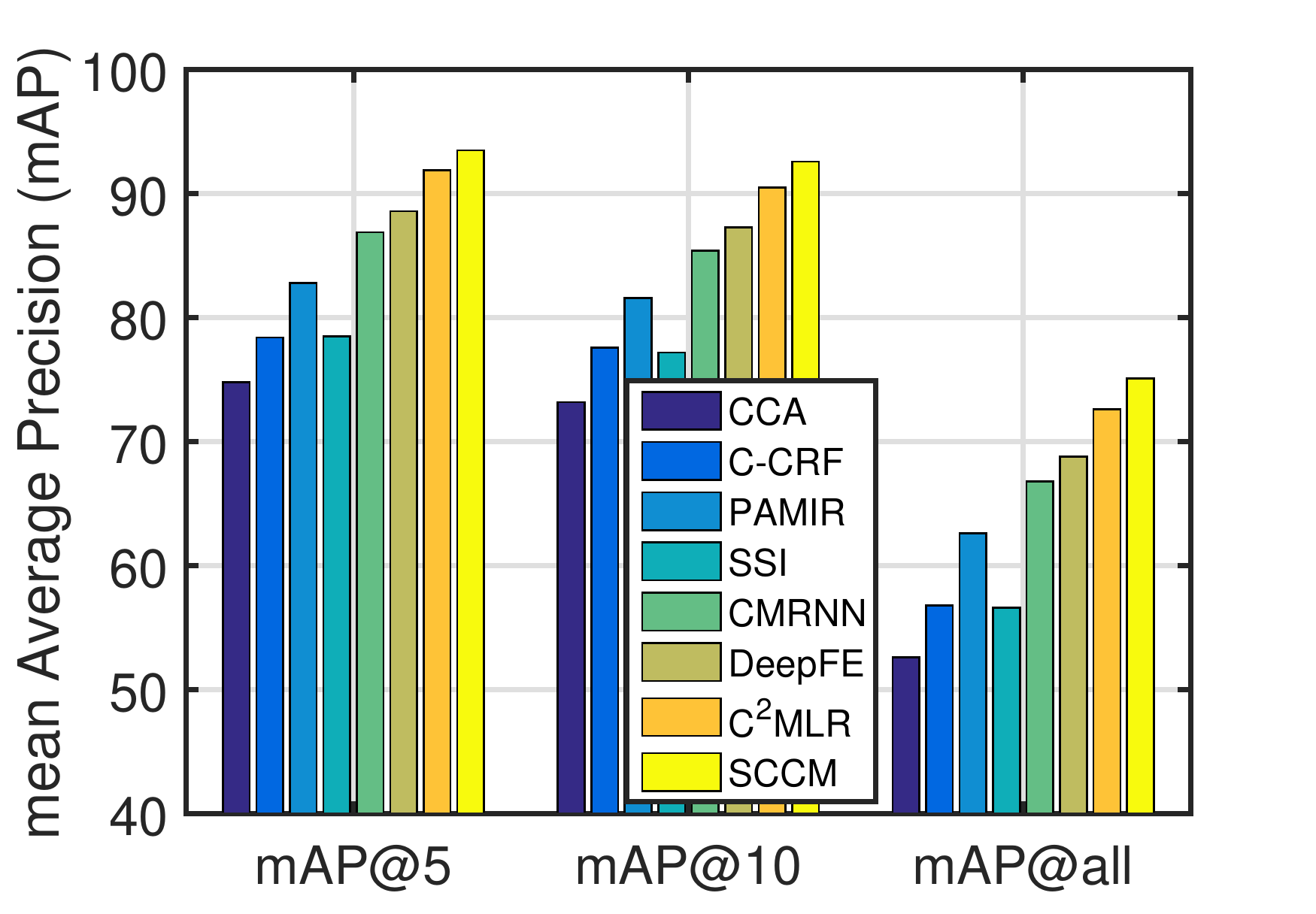}\label{fig:nuswide_text2image}}\\
	\vspace{-0.5em}
	\caption{The performance comparison in terms of mAP over NUS-WIDE dataset. }
	\vspace{-0.5em}
\end{figure}

\begin{figure}[t]
	\centering
	\subfloat[Wiki Image-Query-Text]{\includegraphics[scale=0.23]{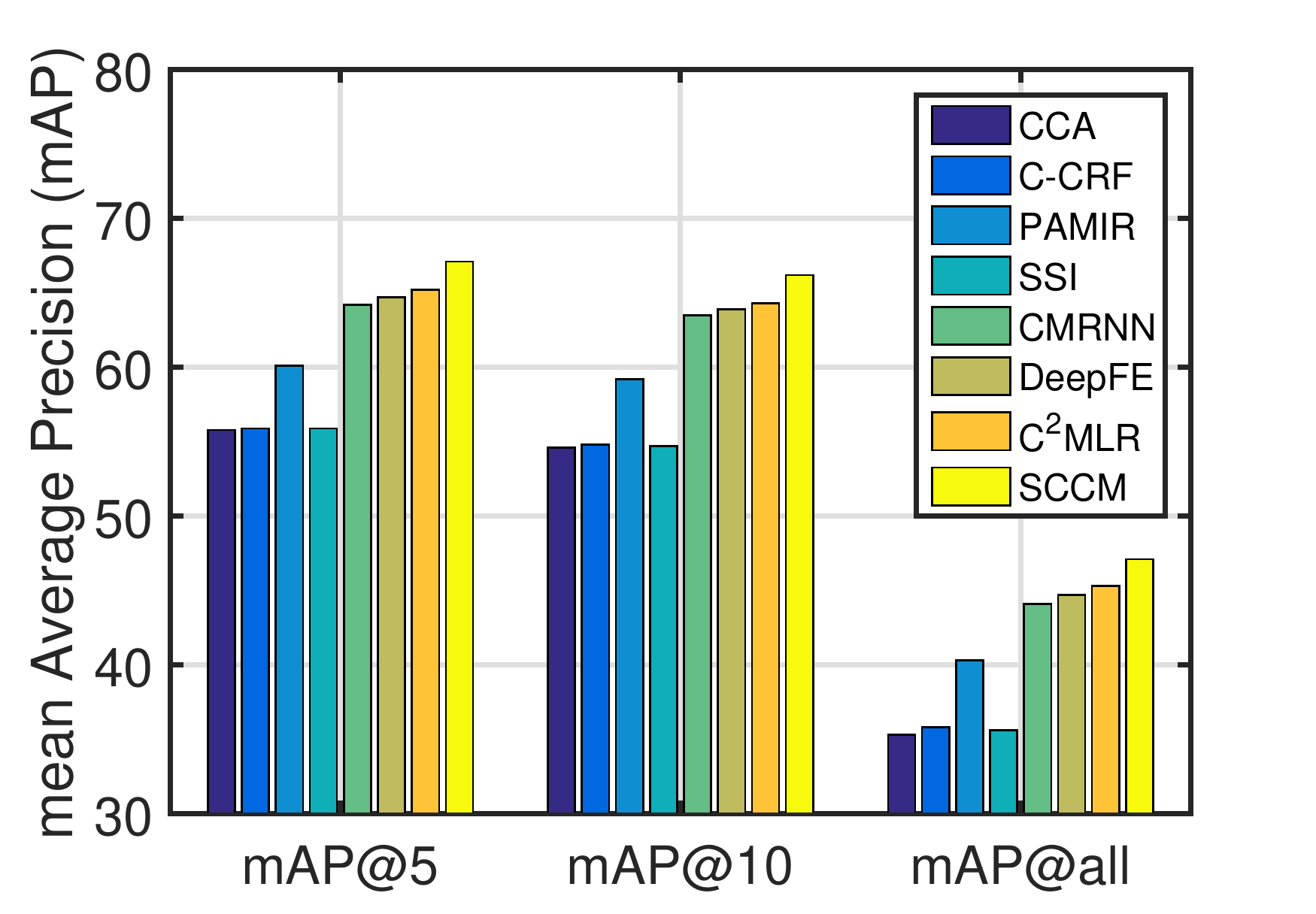}\label{fig:wiki_image2text}} 
	\subfloat[Wiki Text-Query-Image]{\includegraphics[scale=0.23]{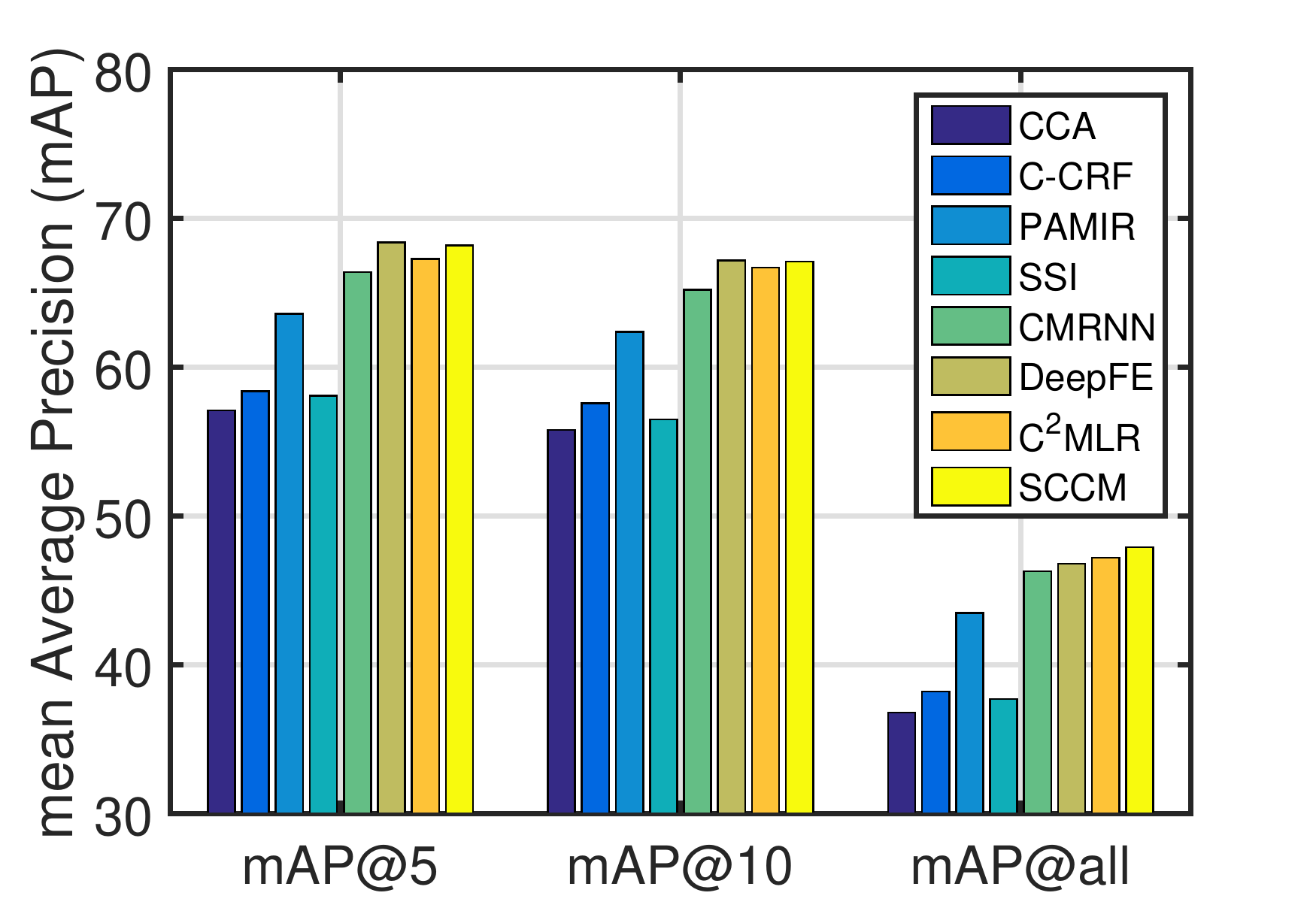}\label{fig:wiki_text2image}}
	\vspace{-0.5em}
	\caption{The performance comparison in terms of mAP over Wiki dataset. }
	\vspace{-0.5em}
\end{figure}

\Cref{fig:nuswide_image2text} (Image-Query-Text) and \Cref{fig:nuswide_text2image} (Text-Query-Image) show the performance comparison of cross-modal ranking in terms of mAP over NUS-WIDE dataset. We have the following observations from the experimental results:

\begin{enumerate}[$\bullet$,leftmargin=*,topsep=0pt,noitemsep]
	\item[--] The proposed \textbf{SCCM} outperforms the other alternatives on both search directions by a large margin in terms of all the evaluation metrics.
	
	\item[--] SSI generally performs better than CCA. Compared to CCA, SSI introduces a nonlinear projection to map the multi-modal data into a common space. This observation verifies that a nonlinear projection for multi-modal learning achieves better performance than a linear projection.
	
	\item[--] The ranking algorithms with local alignment or global alignment generally outperforms the other alternatives. For example, DeepFE and C$^2$MLR achieves significant improvement over CCA, C-CRF, PAMIR and CMRNN. However, these algorithms are built on the assumption that there is an explicit alignment between visual objects and textual words.
\end{enumerate}

We report the experimental results over Wiki dataset in \Cref{fig:wiki_image2text} (Image-Query-Text) and \Cref{fig:wiki_text2image} (Text-Query-Image). By comparing the performance of different alternatives on this dataset, we have the following observations:

\begin{enumerate}[$\bullet$,leftmargin=*,topsep=0pt,noitemsep]
	\item[--] By comparing the performance in the two directions (image-query-text and text-query-image), most of the performance obtain unbalanced performance. In contrast, CCA gets very similar performances in both directions of the retrieval. The reason is that CCA takes strictly paired multi-modal data as the training instances, which makes CCA tend to capture the pair-correspondence between multi-modal data and is unable to capture the discriminative information between multi-modal data.
	
	\item[--] PAMIR gets the best performance among the compared non-deep methods. This phenomenon is because PAMIR maps the images into the textual space while the high-level semantics delivered by the textual space is reasonable enough to get good performance on ranking text. 
\end{enumerate}

Based on the above observations, we conclude that (1) non-linear mapping function contributes to the cross-modal learning to rank algorithm because it is able to capture more sophisticated cross-modal correspondence; (2) it is advantageous to learn an optimal multi-modal embedding space gradually from easy to complex rankings by diverse image queries.

\begin{table}[t]
	\tabcolsep=1 pt
	\center
	\caption{Performance comparison of the proposed algorithm w/t and w/o diversity. Mean average precision (mAP) is used as an evaluation metric. Results are shown in percentages. Larger mAP indicates better performance.}{%
		\begin{tabular}{|c|c|c|c|c|c|}
			\hline
			\multicolumn{2}{|{c}|}{\multirow{2}{*}{Dataset}} & \multicolumn{2}{{c}|}{IQT}    & \multicolumn{2}{{c}|}{{TQI}}\\
			\cline{3-6}
			\multicolumn{2}{|{c}|}{}
			&w/t div.&w/o div.&w/t div.&w/o div.\\ \hline
			\multirow{3}{*}{Pascal'07}				
			&mAP@5	&	\textbf{83.6}	&	79.7	& \textbf{81.8} & 77.4	\\ 
			\cline{2-6}
			&mAP@10	&	\textbf{82.3}	&	78.9	&	\textbf{80.9}	&	76.2	\\ 
			\cline{2-6}
			&mAP@all	&	\textbf{64.1}	&	59.6	&	\textbf{62.6}	&	57.5	\\
			\hline
			\multirow{3}{*}{NUSWIDE}
			&mAP@5	&	\textbf{94.2}	&	90.8	&	\textbf{93.5}	&	89.2	\\ 
			\cline{2-6}
			&mAP@10	&	\textbf{93.6}	&	89.9	&	\textbf{92.6}	&	88.4	\\ 
			\cline{2-6}
			&mAP@all	&	\textbf{74.1}	&	69.7	&	\textbf{75.3}	&	70.5	\\ 
			\hline
			\multirow{3}{*}{Wiki}
			&mAP@5	&	\textbf{67.1}	&	62.3	&	\textbf{68.2}	&	63.6	\\ 
			\cline{2-6}
			&mAP@10	&	\textbf{66.2}	&	61.6	&	\textbf{67.1}	&	62.8	\\ 
			\cline{2-6}
			&mAP@all	&	\textbf{47.1}	&	44.7	&	\textbf{47.9}	&	43.6	\\
			\hline
		\end{tabular}}
		\label{tab:diversity}
	\end{table}

	\subsection{Does Diversity Help?}
	
	In this section, we first conduct an experiment to evaluate whether diversity contribute much to the subsequent performance. By setting $\gamma = 0$ in the optimization problem (\ref{self_obj}), we obtain a modified framework without diversity. Regarding to the task of Image-Query-Text (IQT) and Text-Query-Image (TQI), \Cref{tab:diversity} shows the performance comparison between with diversity (w/t div.) and without diversity (w/o div.). The results clearly demonstrated that the proposed method with diversity consistently outperforms the framework without diversity in both retrieval directions over the three datasets. For example, for the retrieval task of Image-Query-Text, the framework with diversity outperforms the model without diversity, which is 64.1 \emph{v.s.} 59.6 over the Pascal'07 dataset. We attribute this significant improvement to adaptive distinguishing the contributions of varying rankings to the shared space learning and considering the diversity of rankings by different queries. 
	
	\begin{figure}[t]
		\centering
		\subfloat[Pascal'07 Image-Query-Text]{\includegraphics[scale=0.22]{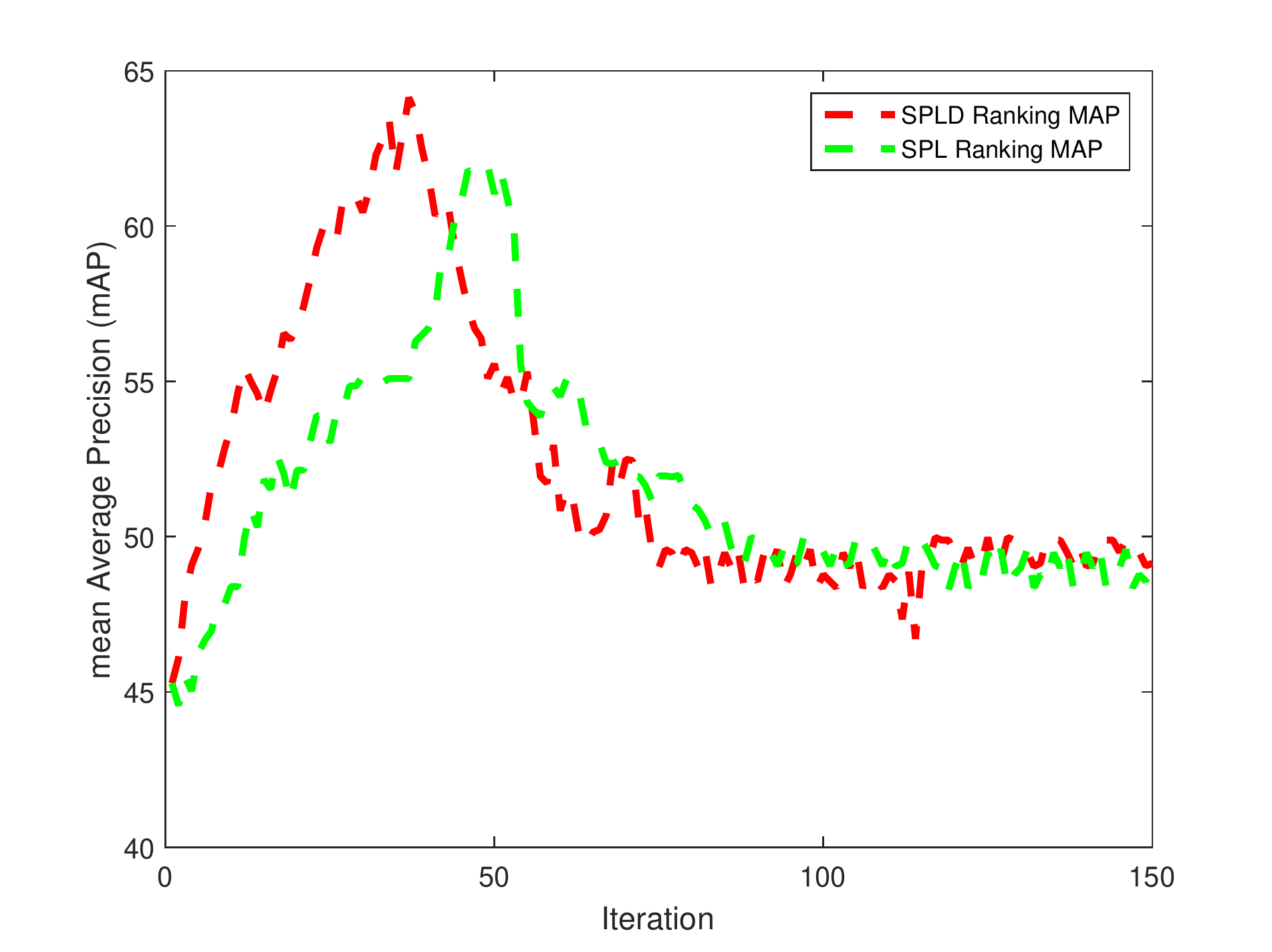}\label{fig:iteration}} 
		\subfloat[Pascal'07 Text-Query-Image]{\includegraphics[scale=0.22]{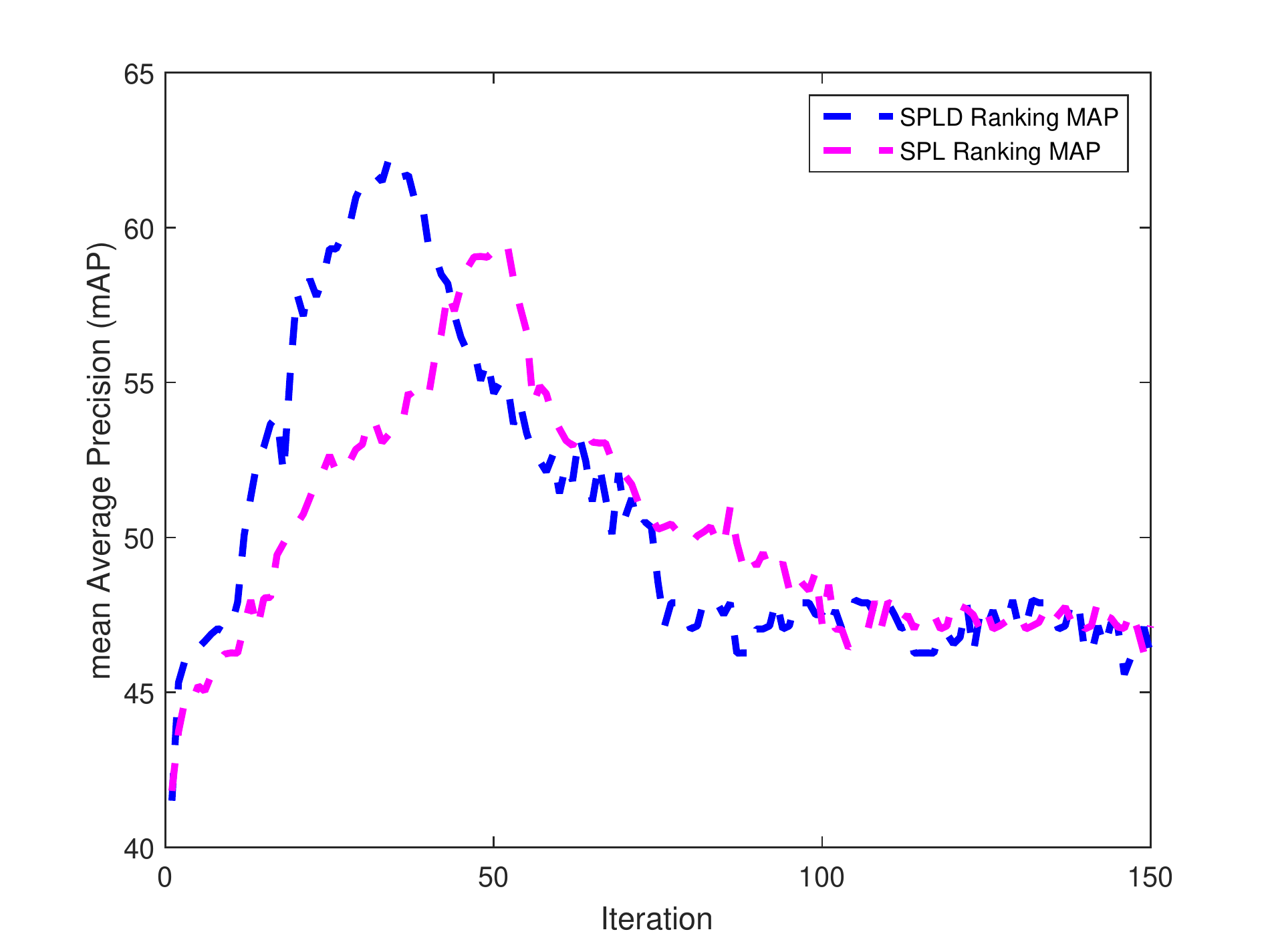}}
		\vspace{-0.8em}	
		\caption{Performance comparison over Pascal'07 dataset between SCCM w/t and w/o diversity \wrt various iterations.}
		\label{fig:validation1}
		\vspace{-0.8em}
		%
		\centering
		\subfloat[NUS-WIDE Image-Query-Text]{\includegraphics[scale=0.22]{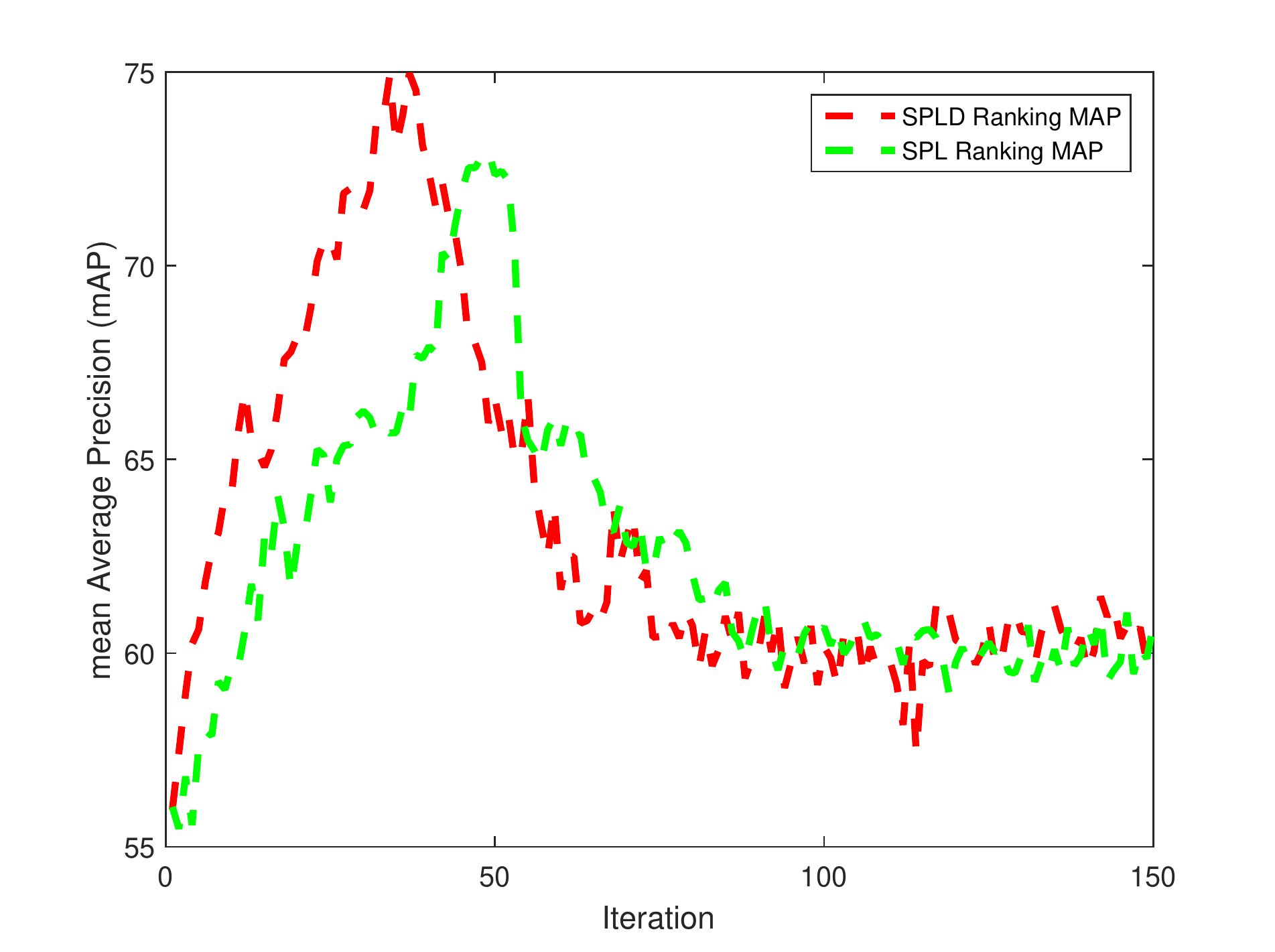}}
		\subfloat[NUS-WIDE Text-Query-Image]{\includegraphics[scale=0.22]{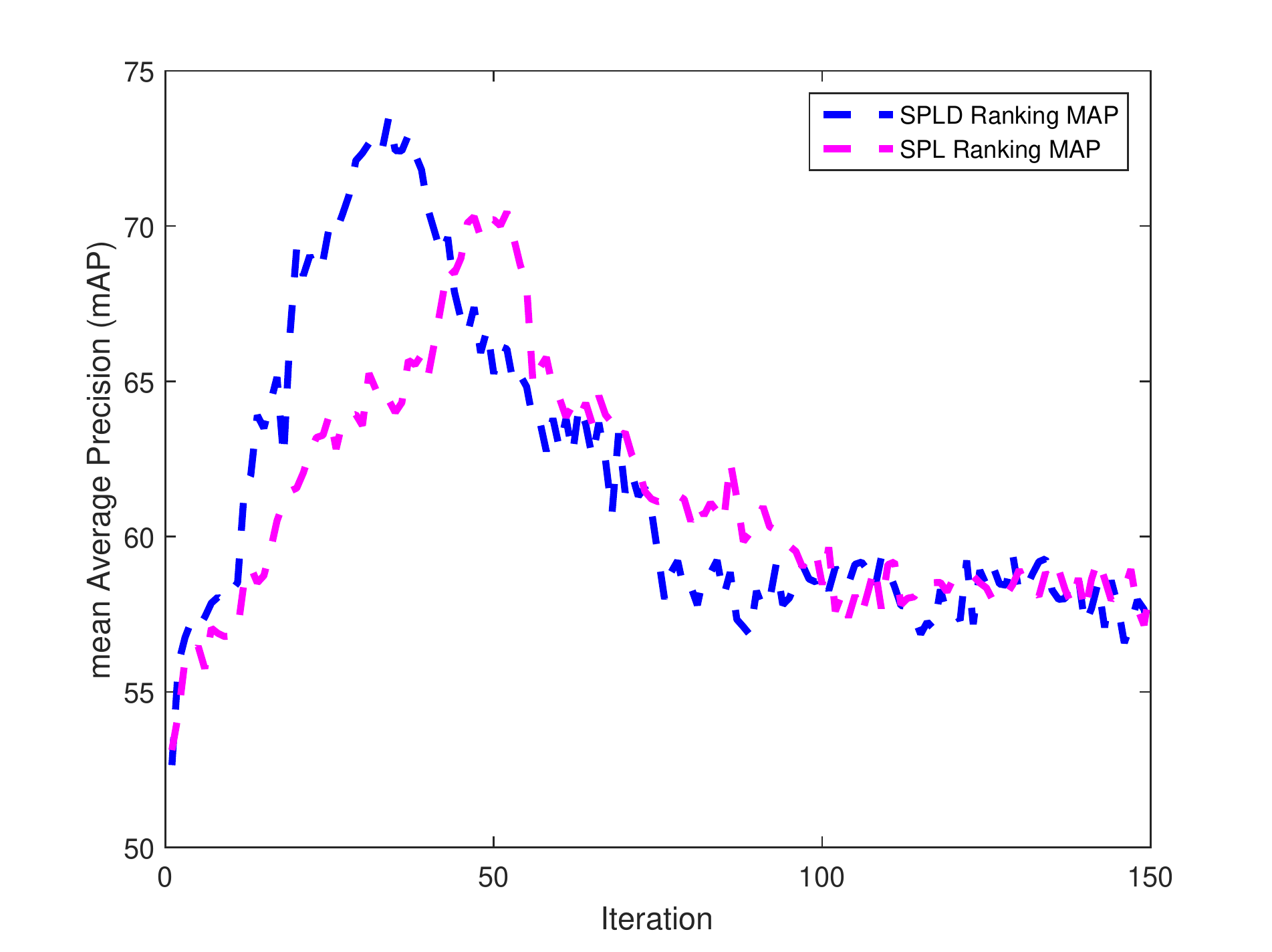}}
		\vspace{-0.8em}
		\caption{Performance comparison over NUS-WIDE dataset between SCCM w/t and w/o diversity \wrt various iterations.}
		\label{fig:validation2}
	\end{figure}
	\begin{figure}[t]
		\centering
		\subfloat[Wiki Image-Query-Text]{\includegraphics[scale=0.22]{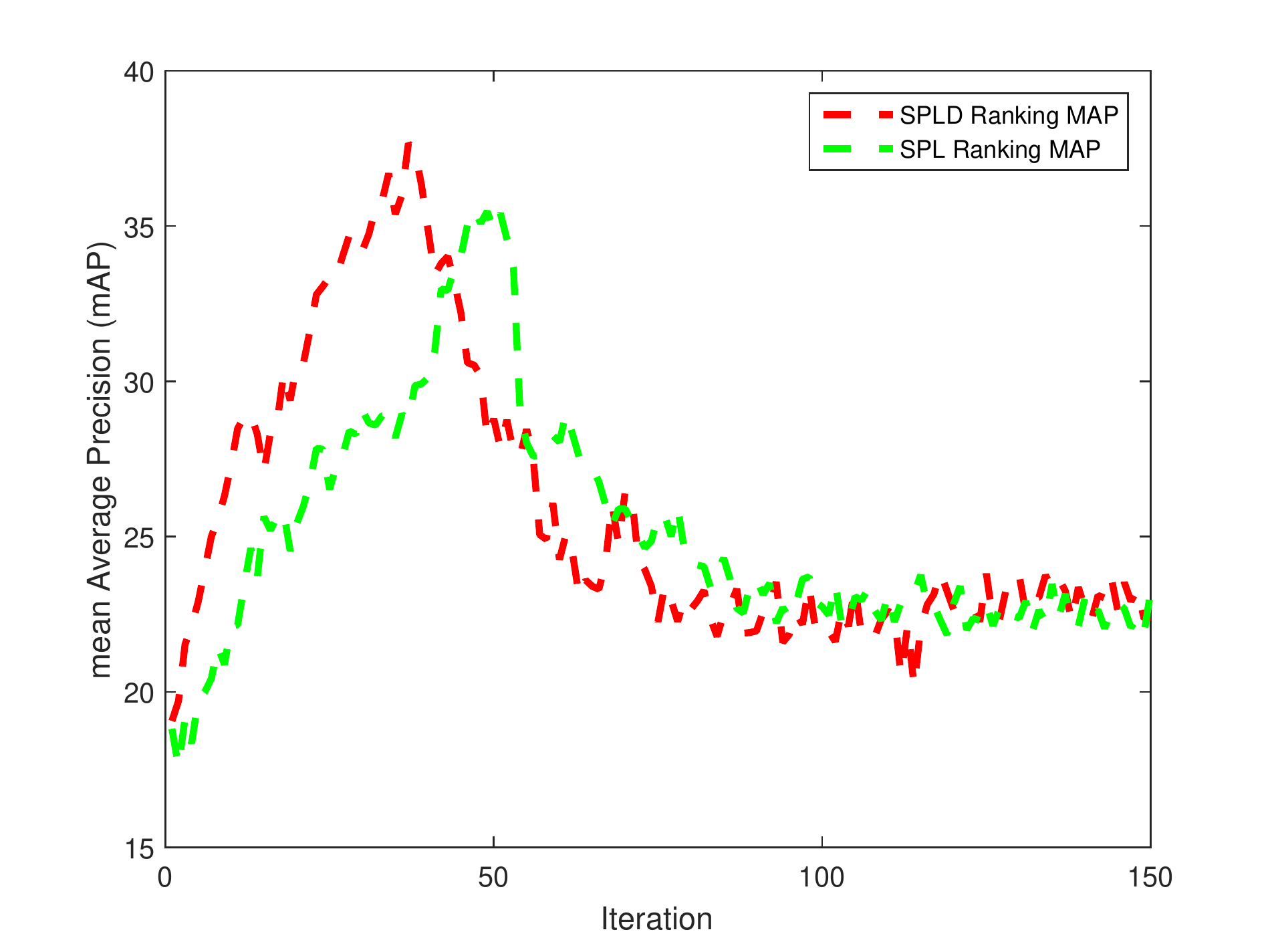}}
		\subfloat[Wiki Text-Query-Image]{\includegraphics[scale=0.22]{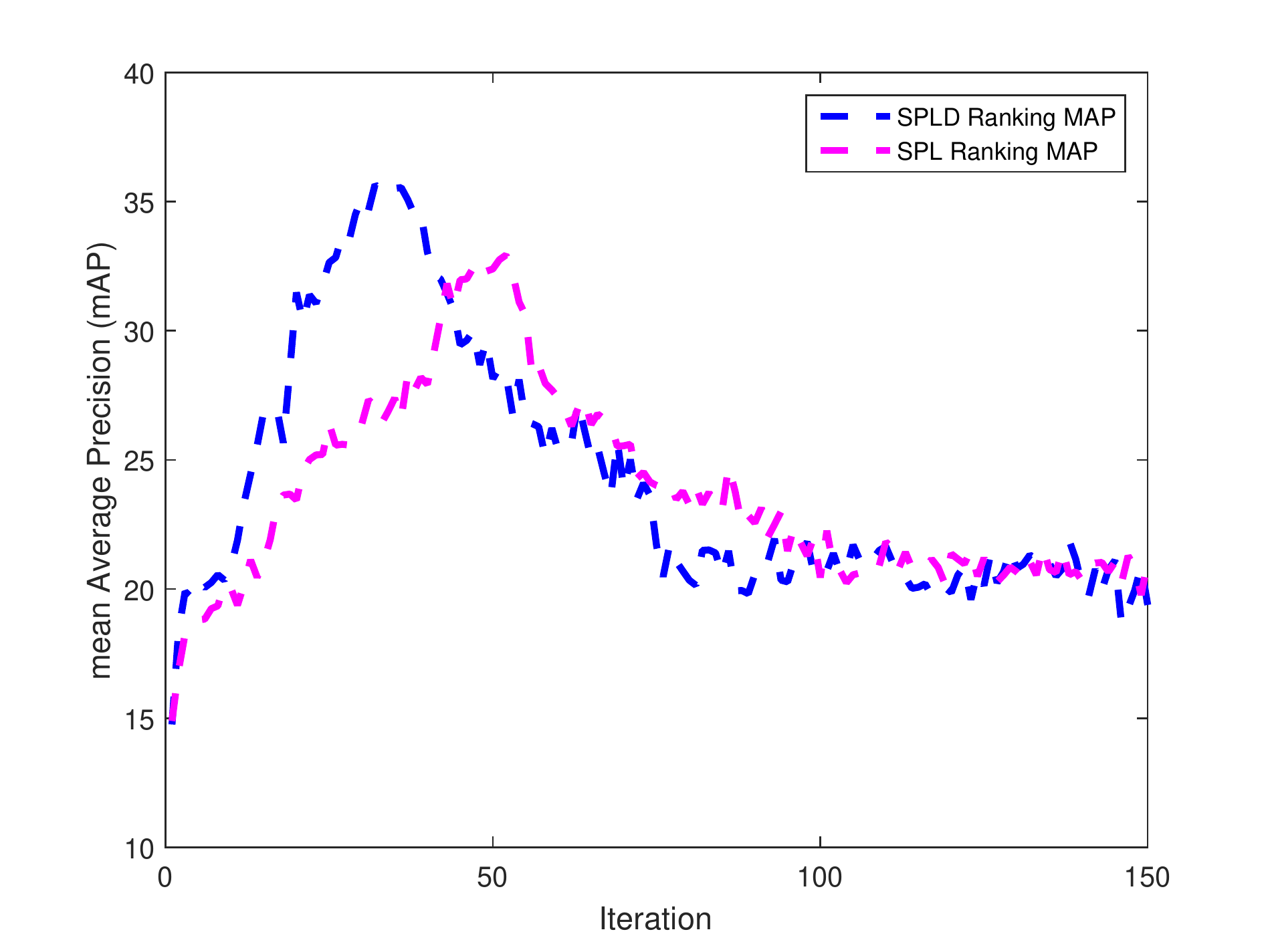}}
		\vspace{-0.8em}
		\caption{Performance comparison over Wiki dataset between SCCM w/t and w/o diversity \wrt various iterations.}
		\label{fig:validation3}
		\vspace{-1.8em}
	\end{figure} 
	
	In addition, we plot performance comparison between SCCM w/t and w/o diversity \wrt various iterations in terms of mAP@all on all three datasets in two retrieval directions in \Cref{fig:validation1} to \Cref{fig:validation3}. A common phenomenon is that the performance improves as the iteration increases. After the performance arrives at its peak, the performance will drop if more iterations are conducted. This is because that some complex image-sentence pairs with large loss values have negative effect on the performance. By comparing SCCM w/t and w/o diversity, we observe that if diversity is considered, SCCM attains a better solution within fewer iterations. For example, in \Cref{fig:iteration}, SCCM w/t diversity obtains the best performance by 36 iterations while SCCM w/o diversity arrives at its peak by 58 iterations. This result indicates that SCCM w/t diversity converges much faster than SCCM w/o diversity.
	
	\section{Conclusion}
	\label{sec:conclusion}
	In this paper, we employ non-linear mapping functions from heterogeneous feature spaces into a shared embedding space and incorporate the SPLD theory into the CMLR to train an optimal multi-modal embedding space gradually from easy rankings by diverse queries to more complex ones. 
	This method adaptively distinguishes the contributions of varying rankings to the shared space learning and explicitly considers the diversity of rankings by different queries at the same time.
	These strategies effectively enhance model's robustness to outliers in a theoretically sound manner and improve its generalization capability with more sophisticated non-linear mapping. The comprehensive experimental results on three benchmark datasets have demonstrated the effectiveness and superiority of the proposed approach on both tasks of text query image and image queried text. We also experimentally illustrate the significant necessary of diversity regularization imposed on importance weight vector for cross-modal retrieval. 
	A possible direction for future work may lie in studying list-wise self-paced CMRL problem based on weak-supervised learning and exploiting the potentials of the proposed model in other applications, such as attribute detection \cite{wang2016category}, face aging \cite{wangrecurrent} and action recognition \cite{wang2016collaborative}.
	
	\section*{Acknowledgement}
	This work was funded by the National Science Foundation (NSF) (No. IIS-1650994, No. IIS-1735591), the National Science Foundation of China (Nos. 61502377, 61532004), and China Postdoctoral Science Foundation (No. 2015M582662).

	{\small
		\bibliographystyle{elsarticle-num-names}
		\bibliography{sigproc}
	}

\end{document}